\documentclass{article}

\usepackage{microtype}
\usepackage{graphicx}
\usepackage{booktabs} 

\usepackage{hyperref}

\newcommand{\set}[1]{\mathcal{#1}}
\providecommand{\sE}{\ensuremath{\set{E}}}
\providecommand{\sR}{\ensuremath{\set{R}}}
\providecommand{\sV}{\ensuremath{\set{V}}}
\providecommand{\sX}{\ensuremath{\set{X}}}
\providecommand{\sY}{\ensuremath{\set{Y}}}

\usepackage[accepted]{icml2023}

\usepackage{amsmath}
\usepackage{amssymb}
\usepackage{mathtools}
\usepackage{amsthm}

\usepackage[capitalize,noabbrev]{cleveref}

\usepackage{xspace}
\usepackage{paralist}
\usepackage{algorithm}
\usepackage{algorithmic}
\usepackage{tabularx}
\usepackage{arydshln}
\usepackage{multirow}
\usepackage{booktabs}
\usepackage{graphicx}
\usepackage{caption}
\usepackage{balance}
\usepackage[normalem]{ulem}
\usepackage{subcaption}

\usepackage{wrapfig}

\newcommand{\ours}[0]{\textsc{InGram}\xspace}

\newcommand{\fb}[0]{FB15K237\xspace}
\newcommand{\nell}[0]{NELL-995\xspace}
\newcommand{\wk}[0]{Wikidata68K\xspace}

\newcommand{\nlz}[0]{NL-0\xspace}
\newcommand{\nla}[0]{NL-25\xspace}
\newcommand{\nlb}[0]{NL-50\xspace}
\newcommand{\nlc}[0]{NL-75\xspace}
\newcommand{\nld}[0]{NL-100\xspace}

\newcommand{\fba}[0]{FB-25\xspace}
\newcommand{\fbb}[0]{FB-50\xspace}
\newcommand{\fbc}[0]{FB-75\xspace}
\newcommand{\fbd}[0]{FB-100\xspace}

\newcommand{\wka}[0]{WK-25\xspace}
\newcommand{\wkb}[0]{WK-50\xspace}
\newcommand{\wkc}[0]{WK-75\xspace}
\newcommand{\wkd}[0]{WK-100\xspace}

\newcommand{\nlg}[0]{NELL-995-v1\xspace}

\newcommand{\grail}[0]{GraIL\xspace}
\newcommand{\compile}[0]{CoMPILE\xspace}
\newcommand{\indigo}[0]{INDIGO\xspace}
\newcommand{\snri}[0]{SNRI\xspace}
\newcommand{\rmpi}[0]{RMPI\xspace}
\newcommand{\cgcn}[0]{CompGCN\xspace}
\newcommand{\npiece}[0]{NodePiece\xspace}
\newcommand{\nelp}[0]{NeuralLP\xspace}
\newcommand{\drum}[0]{DRUM\xspace}
\newcommand{\blp}[0]{BLP\xspace}
\newcommand{\qblp}[0]{QBLP\xspace}
\newcommand{\red}[0]{RED-GNN\xspace}
\newcommand{\nbf}[0]{NBFNet\xspace}
\newcommand{\raild}[0]{RAILD\xspace}

\newcommand{\rotate}[0]{RotatE\xspace}
\newcommand{\conglr}[0]{ConGLR\xspace}
\newcommand{\cbgnn}[0]{CBGNN\xspace}

\newcommand{\pcon}[0]{PathCon\xspace}
\newcommand{\tact}[0]{TACT\xspace}
\newcommand{\distm}[0]{DistMult\xspace}
\newcommand{\angel}[0]{GraphANGEL\xspace}
\newcommand{\mean}[0]{MEAN\xspace}
\newcommand{\lan}[0]{LAN\xspace}

\providecommand{\sE}{\ensuremath{\set{E}}}
\providecommand{\sR}{\ensuremath{\set{R}}}
\providecommand{\sV}{\ensuremath{\set{V}}}
\providecommand{\sN}{\ensuremath{\set{N}}}
\providecommand{\hNi}{\ensuremath{\widehat{\set{N}}_i}}
\providecommand{\sX}{\ensuremath{\set{X}}}
\providecommand{\sY}{\ensuremath{\set{Y}}}
\providecommand{\sF}{\ensuremath{\set{F}}}
\providecommand{\sT}{\ensuremath{\set{T}}}

\renewcommand{\vec}[1]{{\bf{#1}}}
\providecommand{\vh}{\ensuremath{\vec{h}}}

\providecommand{\vx}{\ensuremath{\vec{x}}}
\providecommand{\vy}{\ensuremath{\vec{y}}}
\providecommand{\vz}{\ensuremath{\vec{z}}}
\providecommand{\vb}{\ensuremath{\vec{b}}}

\newcommand{\mat}[1]{\boldsymbol{#1}}
\providecommand{\mA}{\ensuremath{\mat{A}}}

\providecommand{\mD}{\ensuremath{\mat{D}}}
\providecommand{\mE}{\ensuremath{\mat{E}}}

\providecommand{\mW}{\ensuremath{\mat{W}}}
\providecommand{\mTh}{\ensuremath{\mat{P}}} 
\providecommand{\mhTh}{\ensuremath{\widehat{\mat{P}}}} 
\providecommand{\mH}{\ensuremath{\mat{H}}}
\providecommand{\mhH}{\ensuremath{\widehat{\mat{H}}}}
\providecommand{\mM}{\ensuremath{\mat{M}}}
\providecommand{\mhM}{\ensuremath{\widehat{\mat{M}}}}

\providecommand{\mhW}{\ensuremath{\widehat{\mat{W}}}}

\providecommand{\vhx}{\ensuremath{\widehat{\vec{x}}}}
\providecommand{\vhy}{\ensuremath{\widehat{\vec{y}}}}

\providecommand{\mtW}{\ensuremath{\overline{\mat{W}}}}

\DeclareMathOperator{\diag}{diag}
\DeclareMathOperator{\maxf}{max}
\DeclarePairedDelimiter{\ceil}{\lceil}{\rceil}

\theoremstyle{plain}

\theoremstyle{definition}

\theoremstyle{remark}

\usepackage[textsize=tiny]{todonotes}

\icmltitlerunning{\ours: Inductive Knowledge Graph Embedding via Relation Graphs}

\begin{document}

\twocolumn[
\icmltitle{\ours: Inductive Knowledge Graph Embedding via Relation Graphs}



\icmlsetsymbol{equal}{*}

\begin{icmlauthorlist}
\icmlauthor{Jaejun Lee}{sch}
\icmlauthor{Chanyoung Chung}{sch}
\icmlauthor{Joyce Jiyoung Whang}{sch}
\end{icmlauthorlist}

\icmlaffiliation{sch}{School of Computing, KAIST, Daejeon, South Korea}

\icmlcorrespondingauthor{Joyce Jiyoung Whang}{jjwhang@kaist.ac.kr}

\icmlkeywords{Machine Learning, ICML}

\vskip 0.3in
]



\printAffiliationsAndNotice{}  

\begin{abstract}
Inductive knowledge graph completion has been considered as the task of predicting missing triplets between new entities that are not observed during training. While most inductive knowledge graph completion methods assume that all entities can be new, they do not allow new relations to appear at inference time. This restriction prohibits the existing methods from appropriately handling real-world knowledge graphs where new entities accompany new relations. In this paper, we propose an \textsc{In}ductive knowledge \textsc{Gra}ph e\textsc{m}bedding method, \ours, that can generate embeddings of new relations as well as new entities at inference time. Given a knowledge graph, we define a relation graph as a weighted graph consisting of relations and the affinity weights between them. Based on the relation graph and the original knowledge graph, \ours learns how to aggregate neighboring embeddings to generate relation and entity embeddings using an attention mechanism. Experimental results show that \ours outperforms 14 different state-of-the-art methods on varied inductive learning scenarios.
\end{abstract}

\section{Introduction}
Knowledge graphs represent known facts as a set of triplets, each of which is composed of a head entity, a relation, and a tail entity~\cite{survey}. Among various approaches to predicting missing triplets in knowledge graphs, embedding-based methods are known to be effective, where entities and relations are converted into low-dimensional embedding vectors~\cite{kbat}. Classical knowledge graph embedding models~\cite{anal,rotate} assume a transductive learning. That is, it is assumed that all entities and relations are observed during training. Transductive knowledge graph embedding methods predict a missing triplet by identifying a plausible combination of the observed entities and relations~\cite{kgs}. 

\begin{figure}[t]
\centering
\includegraphics[width=\columnwidth]{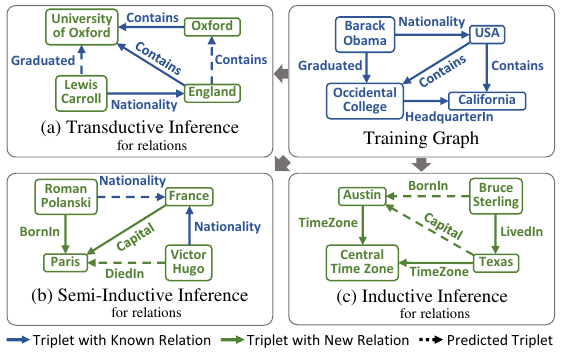}
\caption{For relations, (a) is a transductive inference, (b) is a semi-inductive inference, and (c) is an inductive inference.}
\label{fig:prob}
\end{figure}

\begin{figure*}[t]
\centering
\includegraphics[width=\textwidth]{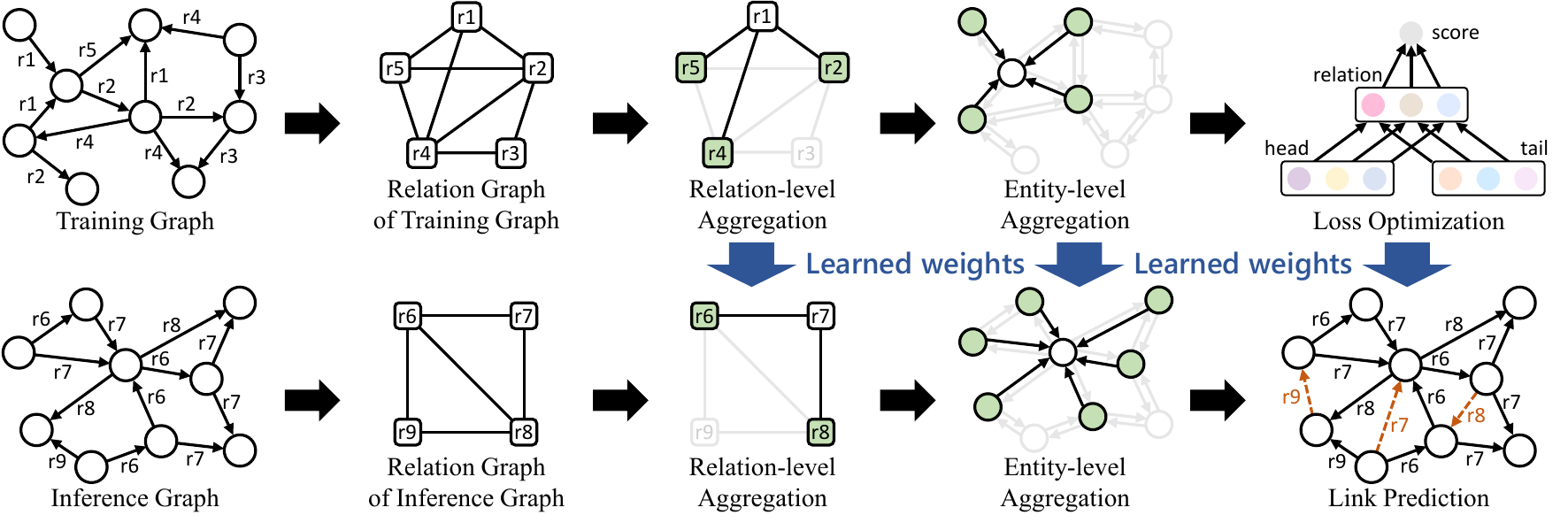}
\caption{Overview of \ours. Given a knowledge graph, a relation graph is created to define the neighboring relations of each relation. Based on the relation graph and the original knowledge graph, relation and entity embedding vectors are computed by aggregating their neighbors' embeddings. During training, \ours learns how to aggregate the neighbors' embeddings by maximizing the scores of training triplets. At inference time, \ours creates embeddings of new relations and entities by aggregating their neighbors' embeddings and conducts link prediction in the way it learned during training.}
\label{fig:oview}
\end{figure*}

In recent years, inductive knowledge graph completion has been studied to predict missing triplets between new entities that are not observed at training time~\cite{grail}. If an entity or a relation is observed during training, we call them \emph{known} and \emph{new} otherwise; they are sometimes referred to as \emph{seen} and \emph{unseen}~\cite{snri}. To handle new entities, some methods focus on learning entity-independent relational patterns by logical rule mining~\cite{drum}, while others exploit Graph Neural Networks (GNNs)~\cite{cbgnn}. However, most existing methods assume that only entities can be new, and all relations should be observed during training. Thus, they perform inductive inference for entities but \emph{transductive} inference for \emph{relations} as shown in Figure~\ref{fig:prob}(a). 

In this paper, we consider more realistic inductive learning scenarios: (i) the relations at inference time consist of a mixture of known and new relations (\emph{semi-inductive} inference for \emph{relations}), or (ii) the relations are all new due to an entirely new set of entities (\emph{inductive} inference for \emph{relations}). Figure~\ref{fig:prob}(b) and  Figure~\ref{fig:prob}(c) show these scenarios. We propose \ours, an \textsc{In}ductive knowledge \textsc{Gra}ph e\textsc{m}bedding method that can generate embedding vectors for new relations and entities only appearing at inference time. Figure~\ref{fig:oview} shows an overview of \ours when all relations and entities are new. A key idea is to define a relation graph where a node corresponds to a relation, and an edge weight indicates the affinity between relations. Once the relation graph is defined, we can designate \emph{neighboring} relations for each relation. Given the relation graph and the original knowledge graph, the relation and entity embedding vectors are computed by attention-based aggregations of their neighbors' embeddings. The aggregation process is optimized to maximize the plausibility scores of triplets in a training knowledge graph. What \ours learns during training is \emph{how to aggregate neighboring embeddings} to generate the relation and entity embeddings. At inference time, \ours generates embeddings of new relations and entities by aggregating neighbors' embeddings based on the new relation graph computed from a given inference knowledge graph and the attention weights learned during training. 

To the best of our knowledge, \ours is the first method that introduces the relation-level aggregation that allows the model to be generalizable to new relations. Due to the fully inductive capability of \ours, we can generate embeddings by training \ours on a tractable, partially observed set and simply applying it to an entirely new set without retraining. Different from some inductive methods that rely on large language models~\cite{bertrl}, \ours makes inferences solely based on the structure of a given knowledge graph. Experimental results show that \ours significantly outperforms 14 different knowledge graph completion methods in inductive link prediction on 12 datasets with varied ratios of new relations. The performance gap between \ours and the best baseline method is substantial, especially when the ratio of new relations is high, which is a more challenging scenario.\footnote{\url{https://github.com/bdi-lab/InGram}}

\begin{figure*}[t]
\centering
\begin{subfigure}{0.31\textwidth}
  \centering
  \includegraphics[height=3.8cm]{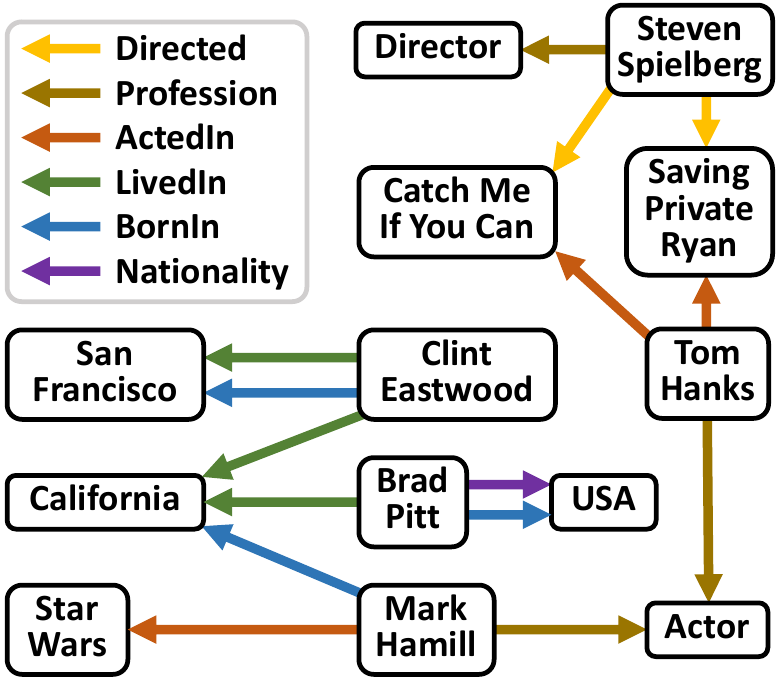}
  \caption{Knowledge Graph}
\end{subfigure}%
\hspace{0.4cm}
\begin{subfigure}{0.24\textwidth}
  \centering
  \includegraphics[height=3.8cm]{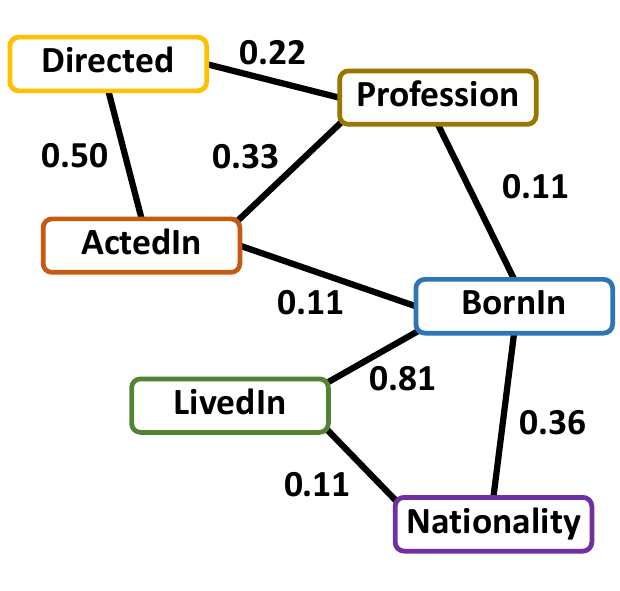}
  \caption{Relation Graph}
\end{subfigure}
\hspace{0.4cm}
\begin{subfigure}{0.36\textwidth}
  \centering
  \includegraphics[height=3.8cm]{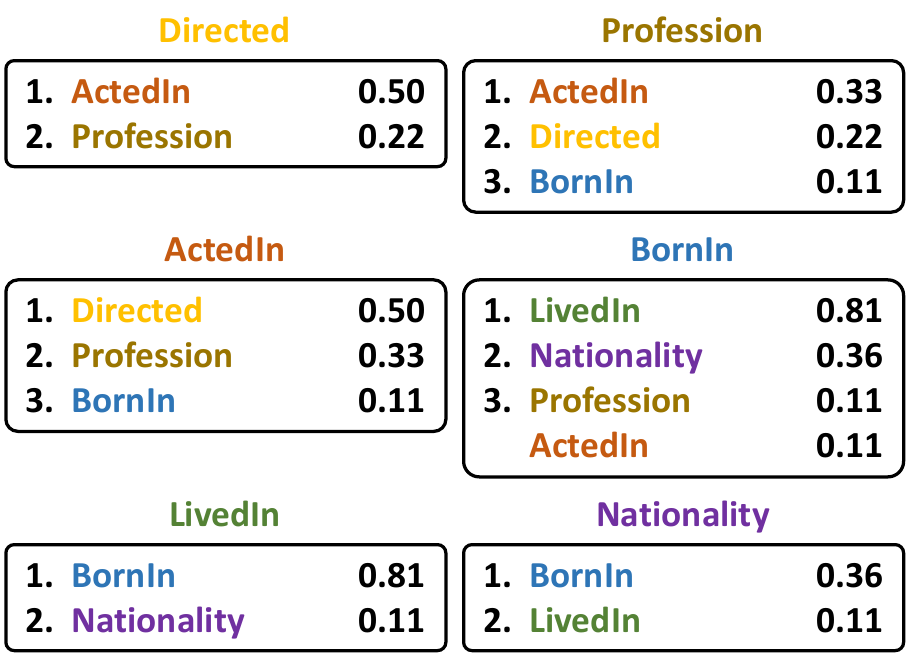}
  \caption{Affinity scores of the relations}
\end{subfigure}%
\caption{Given a knowledge graph, we define a relation graph as a weighted graph where each node indicates a relation, and each edge weight indicates the affinity between two relations. Self-loops in the relation graph are omitted for brevity.}
\label{fig:rgraph}
\end{figure*}

\section{Related Work}
\paragraph*{Rule Mining and Subgraph Reasoning.} For inductive knowledge graph completion,~\cite{nelp} and~\cite{drum} have proposed learning first-order logical rules, while~\cite{pcon},~\cite{nbf} and~\cite{red} have considered relational context or paths. \grail~\cite{grail} has proposed a subgraph-based reasoning framework that extracts subgraphs and scores them using a GNN. Some follow-up works of \grail include \cite{compile},~\cite{snri}, and~\cite{conglr}. \cite{cbgnn} has focused on cycle-based rule learning while~\cite{indigo} has proposed GNN-based encoding capturing logical rules. Different from our method, all these methods assume only entities can be new, and relations should be known in advance. \cite{angel} has recently proposed \angel handling new relations, but it assumes all entities are known. 
\paragraph*{Differences between \rmpi and \ours.} \rmpi~\cite{rmpi} has concurrently studied the problem of handling new relations, although how \rmpi and \ours solve the problem is quite different. While \rmpi extracts a local subgraph for every candidate entity to score the corresponding triplet, \ours directly utilizes the whole structure of a given knowledge graph. Also, \rmpi uses an unweighted relation view per every individual triplet, whereas \ours defines one global relation graph where weights are important. Due to these fundamental differences, \ours is much more scalable and effective than \rmpi. For example, \ours took 15 minutes and \rmpi took 52 hours to process \nld dataset while \ours achieves much better link prediction performance than \rmpi. Details will be discussed in Section~\ref{sec:exp}. Furthermore, \rmpi is designed only for knowledge graph completion and does not compute embedding vectors, while \ours returns a set of embedding vectors for entities and relations that can be also utilized in many other downstream tasks. 
\paragraph*{Reasoning on Evolving Graphs.} Some methods have focused on modeling emerging entities~\cite{mean,cfag}. For example,~\cite{lan} has considered rule and network-based attention weights, and~\cite{invt} has extended~\rotate~\cite{rotate}. All these methods assume that a triplet should be composed of one known entity and one new entity.~\cite{argcn} has proposed a GCN-based~\cite{gcn} method to more flexibly handle emerging entities. Compared to these methods, we tackle a more challenging problem where all entities are new instead of only some portions being new.
\paragraph*{Textual Descriptions and Language Models.}~\cite{blp},~\cite{qblp} and~\cite{raild} have used pre-trained vectors by BERT~\cite{bert} using textual descriptions. Some methods have utilized language models to handle new entities~\cite{statik}. For example,~\cite{bertrl} has leveraged a pre-trained language model and fine-tuned it. All these methods employ a rich language model or require text descriptions that might not always be available. On the other hand, \ours only utilizes the structure of a given knowledge graph.


\section{Problem Definition and Setting}
\label{sec:problem}
In inductive knowledge graph embedding, we are given two graphs: a training graph $\widetilde{G_{\text{tr}}}$ and an inference graph $\widetilde{G_{\text{inf}}}$. A training graph is defined by $\widetilde{G_{\text{tr}}}=(\sV_{\text{tr}},\sR_{\text{tr}},\sE_{\text{tr}})$ where $\sV_{\text{tr}}$ is a set of entities, $\sR_{\text{tr}}$ is a set of relations, and $\sE_{\text{tr}}$ is a set of triplets in $\widetilde{G_{\text{tr}}}$. We divide $\sE_{\text{tr}}$ into two disjoint sets such as $\sE_{\text{tr}}\coloneqq \sF_{\text{tr}} \cup \sT_{\text{tr}}$ where $\sF_{\text{tr}}$ is a set of known facts and $\sT_{\text{tr}}$ is a set of triplets a model is optimized to predict. An inference graph is defined by $\widetilde{G_{\text{inf}}}=(\sV_{\text{inf}},\sR_{\text{inf}},\sE_{\text{inf}})$ where $\sV_{\text{inf}}$ is a set of entities, $\sR_{\text{inf}}$ is a set of relations, and $\sE_{\text{inf}}$ is a set of triplets in $\widetilde{G_{\text{inf}}}$. We partition $\sE_{\text{inf}}$ into three pairwise \emph{disjoint} sets, such that $\sE_{\text{inf}}\coloneqq \sF_{\text{inf}} \cup \sT_{\text{val}} \cup \sT_{\text{test}}$ with a ratio of 3:1:1. $\sF_{\text{inf}}$ is a set of observed facts where it contains all entities and relations included in $\widetilde{G_{\text{inf}}}$, $\sT_{\text{val}}$ is a set of triplets for validation, and $\sT_{\text{test}}$ is a set of test triplets. The fact sets $\sF_{\text{tr}}$ and $\sF_{\text{inf}}$ are also defined in~\cite{red,qblp}. Note that $\sV_{\text{tr}} \cap \sV_{\text{inf}}=\emptyset$ by following a conventional inductive learning setting~\cite{grail}. Most existing methods assume $\sR_{\text{inf}} \subseteq \sR_{\text{tr}}$ due to the constraint that relations cannot be new at inference time~\cite{red}. On the other hand, in our problem setting, $\sR_{\text{inf}}$ is not necessarily a subset of $\sR_{\text{tr}}$ since new relations are allowed to appear.

At training time, we use $G_{\text{tr}}\coloneqq(\sV_{\text{tr}},\sR_{\text{tr}},\sF_{\text{tr}})$ and a model is trained to predict $\sT_{\text{tr}}$. When tuning the hyperparameters of a model, we use $G_{\text{inf}}\coloneqq(\sV_{\text{inf}},\sR_{\text{inf}},\sF_{\text{inf}})$ to compute the embeddings; we check the model's performance on $\sT_{\text{val}}$. At inference time, we evaluate the model's performance using $\sT_{\text{test}}$. The way we use $\sF_{\text{inf}}$, $\sT_{\text{val}}$ and $\sT_{\text{test}}$ is identical to~\cite{qblp,npiece}. For brevity, we do not explicitly mention $G_{\text{tr}}$ or $G_{\text{inf}}$ in the following sections; those should be distinguished in context.

\section{Defining Relation Graphs}
\label{sec:rgraph}
Let us represent a knowledge graph as $G=(\sV, \sR, \sF)$ where $\sV$ is a set of entities, $\sR$ is a set of relations, and $\sF$ is a set of triplets, i.e., $\sF=\{(v_i, r_k, v_j)| v_i\in\sV, r_k\in\sR, v_j\in\sV \}$. For every $(v_i, r_k, v_j)\in\sF$, we add a reverse relation $r_k^{-1}$ to $\sR$ and add a reverse triplet $(v_j, r_k^{-1}, v_i)$ to $\sF$~\cite{rgcn}. Assume that $|\sV|=n$ and $|\sR|=m$. 

Given a knowledge graph, we define a relation graph as a weighted graph where each node corresponds to a relation, and each edge weight indicates the affinity between two relations. Figure~\ref{fig:rgraph} shows an example where we omit the reverse relations and self-loops in the relation graph for simplicity. Briefly speaking, we measure the affinity between two relations by considering how many entities are shared between them and how frequently they share the same entity.

To represent which relations are associated with which entities, we create two matrices $\mE_{\mathrm{h}}\in\mathbb{R}^{n\times m}$ and $\mE_{\mathrm{t}}\in\mathbb{R}^{n\times m}$ where the subscripts $\mathrm{h}$ and $\mathrm{t}$ indicate head and tail, respectively. Let $\mE_{\mathrm{h}}[i,j]$ denote the $i$-th row and the $j$-th column element of $\mE_{\mathrm{h}}$, where $\mE_{\mathrm{h}}[i,j]$ is the frequency of $v_i$ appearing as a head entity of relation $r_j$. Similarly, $\mE_{\mathrm{t}}[i,j]$ is the frequency of $v_i$ appearing as a tail entity of relation $r_j$. While some entities are frequently involved in relations, some entities are rarely involved in relations. To take this into account, we define the degree of an entity to be the sum of its frequencies. Formally, we define $\mA_\mathrm{h}\coloneqq{\mE_{\mathrm{h}}}^T {\mD_\mathrm{h}}^{-2} \mE_{\mathrm{h}}$ where ${\mD_\mathrm{h}}\in\mathbb{R}^{n\times n}$ is the degree diagonal matrix of entities for head, i.e., $\mD_\mathrm{h}[i,i]\coloneqq \sum_j \mE_{\mathrm{h}}[i,j]$. Similarly, we define $\mA_\mathrm{t}\coloneqq{\mE_{\mathrm{t}}}^T {\mD_\mathrm{t}}^{-2} \mE_{\mathrm{t}}$ where ${\mD_\mathrm{t}}\in\mathbb{R}^{n\times n}$ is the degree diagonal matrix of entities for tail. The degree normalization terms allow the sum of the affinity weights introduced by each entity in $\mA_{\mathrm{h}}$ and $\mA_{\mathrm{t}}$ to be normalized to one.

Finally, we define the adjacency matrix of the relation graph to be $\mA\coloneqq \mA_\mathrm{h} + \mA_\mathrm{t}$ where $\mA\in\mathbb{R}^{m\times m}$ and each element $a_{ij}\in\mA$ indicates the affinity between the relations~$r_i$~and~$r_j$. In Figure~\ref{fig:rgraph}, we see that the relation graph identifies semantically close relations even though we use only the structure of a knowledge graph. However, there is a chance that we miss some semantically similar relation pairs in the relation graph if they do not share an entity in the knowledge graph. Note that the goal of the relation graph is not to identify the perfect set of similar relations but to define a relation's reasonable neighborhood whose representation vectors can be used to create the embedding of the target relation, which will be discussed in Section~\ref{sec:rel}. 

\section{\ours: Inductive Knowledge Graph Embedding Model}
\label{sec:method}
We present \ours that consists of relation-level aggregation, entity-level aggregation, and modeling of relation-entity interactions. 

\subsection{Updating Relation Representation Vectors via Relation-Graph-Based Aggregation}
\label{sec:rel}
Suppose we have an initial feature vector for a relation $r_i$, denoted by $\vx_i\in\mathbb{R}^{d}$ ($i=1,{\cdots},m$), where $d$ is the dimension of a relation vector. We initialize $\vx_i$ using Glorot initialization~\cite{glor}. Let $\vz_i^{(l)}\in\mathbb{R}^{d'}$ denote a hidden representation of $r_i$ where $d'$ is the hidden dimension, the superscript $(l)$ indicates the $l$-th layer with $l=0,{\cdots},L-1$, and $L$ is the number of layers for relations. We compute $\vz_i^{(0)} = \mH \vx_i$ where $\mH\in\mathbb{R}^{d' \times d}$ is a trainable matrix that projects the initial feature vector to a hidden representation vector. All vectors are assumed to be column vectors unless specified.

Since we define the relation graph $\mA$ in Section~\ref{sec:rgraph}, we can designate the neighboring relations of each relation using $\mA$. We update each relation's representation by aggregating its own and neighbors' representation vectors. Specifically, we define the forward propagation as follows:

{\small \begin{equation}
\label{eq:relup}
\vz_i^{(l+1)} = \sigma\left(\sum_{r_j\in\sN_i} \alpha_{ij}^{(l)} \mW^{(l)} \vz_j^{(l)}\right)
\end{equation}}where $\sigma(\cdot)$ is an element-wise activation function such as LeakyReLU~\cite{leaky}, $\vz_j^{(l)}$ is a relation representation vector of $r_j$, $\mW^{(l)}\in\mathbb{R}^{{d'}\times{d'}}$ is a weight matrix, $\sN_i$ is the set of neighbors of $r_i$ on the relation graph $\mA$, and the attention value $\alpha_{ij}^{(l)}$ is defined by\footnote{Note that $\sN_i$ includes $r_i$ itself because $\mA$ contains self-loops.} 

{\small \begin{equation}
\label{eq:relatt}
\alpha_{ij}^{(l)} = \dfrac{\textup{exp}\left({\vy}^{(l)}\sigma\left(\mTh^{(l)}[\vz_i^{(l)}\| \vz_j^{(l)}]\right)+c_{s(i,j)}^{(l)}\right)}{\sum_{r_{j'}\in\sN_i} \textup{exp}\left({\vy}^{(l)}\sigma\left(\mTh^{(l)}[\vz_{i}^{(l)}\|\vz_{j'}^{(l)}]\right)+c_{s(i,j')}^{(l)}\right)}   
\end{equation}}where $\|$ denotes concatenating vectors vertically, $\mTh^{(l)}\in\mathbb{R}^{{d'}\times{2d'}}$ is a weight matrix, and $\vy^{(l)}\in\mathbb{R}^{1\times d'}$ is a row weight vector. We apply ${\vy}^{(l)}$ after $\sigma(\cdot)$ to resolve the static attention issue of~\cite{gat} by following~\cite{gat2}. In our implementation, we use the residual connection~\cite{resi} and the multi-head attention mechanism with $K$ heads~\cite{att}. In (\ref{eq:relatt}), $c_{s(i,j)}^{(l)}$ is a learnable parameter indexed by $s(i,j)$ which is defined by

{\small \begin{equation}
\label{eq:indx}
s(i,j)=\ceil*{\dfrac{\text{rank}(a_{ij}) \times B}{\text{nnz}(\mA)}}
\end{equation}}where $a_{ij}$ indicates the value corresponding to the $i$-th row and the $j$-th column in $\mA$, $\text{rank}(a_{ij})$ is the ranking of $a_{ij}$ when the non-zero elements in $\mA$ are sorted in descending order, $\text{nnz}(\mA)$ is the number of non-zero elements in $\mA$, and $B$ is a hyperparameter indicating the number of bins. We divide the relation pairs into $B$ different bins according to their affinity scores, i.e., the $a_{ij}$ values. Each relation pair has an index value of $1\leq s(i,j)\leq B$ and we have the learnable parameters $c_1^{(l)},\cdots,c_B^{(l)}$. 

In (\ref{eq:relup}) and (\ref{eq:relatt}), we update the relation representation vectors by using the attention mechanism, where we consider the relative importance of each neighboring relation and the affinity between the relations. While the former term is computed based on the local structure of the target relation, the latter term, $c_{s(i,j)}^{(l)}$, reflects a global level of affinity because we divide the affinity scores into $B$ different levels globally. When representing a relation's representation vector, it would be beneficial to take the vectors of similar relations to the target relation. Thus, $c_{s(i,j)}^{(l)}$ is expected to have a high value for a small $s(i,j)$ because the relation pairs belonging to a small $s(i,j)$ indicate those having high affinity values. We empirically observed that $c_{s(i,j)}^{(l)}$ values are learned as expected (details in Section~\ref{sec:qual}). The way we incorporate the affinity into the attention mechanism is inspired by Graphormer~\cite{gmer} even though it is not designed for inductive knowledge graph embedding. 

By updating $\vz_i^{(l)}$ for $l=0,{\cdots},L-1$ using (\ref{eq:relup}), we have the final-level relation representation vectors $\vz_i^{(L)}$ for $i=1,{\cdots},m$ which are utilized to update entity representation vectors.

\subsection{Entity Representation by Entity-level Aggregation}
\label{sec:node}
Suppose we have an initial feature vector for an entity $v_i$, denoted by $\vhx_i\in\mathbb{R}^{\widehat{d}}$ ($i=1,{\cdots},n$), where $\widehat{d}$ is the dimension of an entity vector. We initialize $\vhx_i$ using Glorot initialization. Let $\vh_i^{(l)}\in\mathbb{R}^{\widehat{d}'}$ denote a hidden representation of $v_i$ where $\widehat{d}'$ is the hidden dimension, the superscript $(l)$ indicates the $l$-th layer with $l=0,{\cdots},\widehat{L}-1$, and $\widehat{L}$ is the number of layers for entities. We compute $\vh_i^{(0)} = \mhH \vhx_i$ where $\mhH\in\mathbb{R}^{\widehat{d}' \times \widehat{d}}$ is trainable. 

We update a representation vector of $v_i$ by aggregating the representation vectors of its neighbors, its own vector, and the representation vectors of the relations adjacent to $v_i$. When we refer to the relation representation vectors here, we always use the final-level relation representation vectors $\vz_k^{(L)}$ for $k=1,{\cdots},m$ acquired in Section~\ref{sec:rel}. 

We define the neighbors of $v_i\in\sV$ to be $\hNi=\{v_j|(v_j,r_k,v_i)\in\sF, v_j\in\sV, r_k\in\sR\}$. To compute the attention weight for the self-loop of $v_i$, we consider the mean vector of the representation vectors of the relations adjacent to $v_i$: 
\begin{displaymath}
\text{\small$\displaystyle
\label{eq:zi}
\bar{\vz}_i^{(L)} = \sum_{v_j\in\hNi}\sum_{r_k\in{\sR_{ji}}} \dfrac{\vz_{k}^{(L)}}{\sum_{v_{j'}\in\hNi}|\sR_{j'i}|}$%
}
\end{displaymath}where $\sR_{ji}$ denotes the set of relations from $v_{j}$ to $v_i$. We update an entity representation vector of $v_i$ by
{\scriptsize \begin{equation}
\begin{split}
\label{eq:nodeup}
& \vh_i^{(l+1)} = \\
& \sigma\left(\beta_{ii}^{(l)} \mhW^{(l)} [ \vh_{i}^{(l)} \| \bar{\vz}_i^{(L)}] + \sum_{v_j\in\hNi}\sum_{r_k\in{\sR_{ji}}} \beta_{ijk}^{(l)} \mhW^{(l)} [ \vh_{j}^{(l)} \| \vz_{k}^{(L)}]\right)
\end{split}
\end{equation}}where $\mhW^{(l)}\in\mathbb{R}^{\widehat{d}'\times(\widehat{d}'+d')}$ is a weight matrix, and $\beta_{ii}^{(l)}$ and $\beta_{ijk}^{(l)}$ are the attention coefficients which are defined by
{\scriptsize
\begin{displaymath}
\begin{split}
\label{eq:nodeatt1}
& \beta_{ii}^{(l)} = \\
& \dfrac{\textup{exp}\left({\vhy}^{(l)}\sigma\!\left( \mhTh^{(l)} \vb_{ii}^{(l)}\!\right)\right)}{\textup{exp}\!\left({\vhy}^{(l)}\sigma\!\left(\mhTh^{(l)} \vb_{ii}^{(l)}\!\right)\right) + \sum\limits_{v_{j'}\in\hNi} \sum\limits_{r_{k'}\in\sR_{j'i}} \textup{exp}\!\left({\vhy}^{(l)}\sigma\!\left(\mhTh^{(l)} \vb_{ij'k'}^{(l)}\!\right)\right)},
\end{split}
\end{displaymath}
\begin{displaymath}
\begin{split}
\label{eq:nodeatt2}
& \beta_{ijk}^{(l)} = \\
& \dfrac{\textup{exp}\!\left({\vhy}^{(l)}\sigma\!\left(\mhTh^{(l)} \vb_{ijk}^{(l)}\right)\right)}{\textup{exp}\!\left({\vhy}^{(l)}\sigma\!\left(\mhTh^{(l)} \vb_{ii}^{(l)}\right)\right) + \sum\limits_{v_{j'}\in\hNi} \sum\limits_{r_{k'}\in\sR_{j'i}} \textup{exp}\!\left({\vhy}^{(l)}\sigma\!\left(\mhTh^{(l)} \vb_{ij'k'}^{(l)}\right)\right)}
\end{split}
\end{displaymath}}where $\vb_{ii}^{(l)}=[\vh_i^{(l)} \| \vh_i^{(l)} \| \bar{\vz}_i^{(L)}]$, $\vb_{ijk}^{(l)}=[\vh_i^{(l)} \| \vh_j^{(l)} \| \vz_k^{(L)}]$, $\mhTh^{(l)}\in\mathbb{R}^{{\widehat{d}'}\times{(2\widehat{d}'+d')}}$ is a linear transformation matrix, and ${\vhy}^{(l)}\in\mathbb{R}^{1\times \widehat{d}'}$ is a row weight vector. We also implement the residual connection and the multi-heads with $\widehat{K}$ heads. Our formulation in (\ref{eq:nodeup}) seamlessly extends GATv2~\cite{gat2} by incorporating the relation representation vectors in every aggregation step. Specifically, when we regard all relation vectors as constant, (\ref{eq:nodeup}) is equivalent to GATv2. By updating $\vh_i^{(l)}$ for $l=0,{\cdots},\widehat{L}-1$ using (\ref{eq:nodeup}), we have the final-level entity representation vectors $\vh_i^{(\widehat{L})}$ ($i=1,{\cdots},n$) which are utilized to model relation-entity interactions.

\subsection{Modeling Relation-Entity Interactions}
Given the representation vectors provided in Section~\ref{sec:rel} and Section~\ref{sec:node}, we compute the final embedding vectors: $\vz_k \coloneqq \mM \vz_k^{(L)}$ ($k=1,{\cdots},m$) for relations and $\vh_i \coloneqq \mhM \vh_i^{(\widehat{L})}$ ($i=1,{\cdots},n$) for entities, where $\mM\in\mathbb{R}^{d \times d'}$ and $\mhM\in\mathbb{R}^{\widehat{d} \times \widehat{d}'}$ are trainable projection matrices. 

A knowledge graph embedding scoring function, denoted by $f(v_i, r_k, v_j)$, returns a scalar value representing the plausibility of a given triplet $(v_i, r_k, v_j)$~\cite{rgcn}. To model the interactions between relation and entity embeddings, we use a variant of \distm~\cite{distm}. We define our scoring function by 
\begin{equation}
\label{eq:score}
f(v_i, r_k, v_j)\coloneqq \vh_i^T \diag(\mtW\vz_k) \vh_j 
\end{equation}
where $\mtW\in\mathbb{R}^{\widehat{d} \times d}$ is a weight matrix that converts the dimension of $\vz_k$ from $d$ to $\widehat{d}$ and $\diag(\mtW\vz_k)$ is the diagonal matrix whose diagonal is defined by $\mtW\vz_k$. Let $(v_i,r_k,v_j)\in\sT_{\text{tr}}$ be a positive triplet in a training set $\sT_{\text{tr}}$ described in Section~\ref{sec:problem}. We create negative triplets by corrupting a head or a tail entity of a positive triplet. Let $(\mathring{v_i},r_k,\mathring{v_j})\in\mathring{\sT_{\text{tr}}}$ denote the negative triplets. The margin-based ranking loss is defined by

{\small
\begin{displaymath}
\sum_{(v_i,r_k,v_j)\in\sT_{\text{tr}}} \sum_{(\mathring{v_i},r_k,\mathring{v_j})\in\mathring{\sT_{\text{tr}}}} \maxf(0,\gamma-f(v_i, r_k, v_j)+f(\mathring{v_i}, r_k, \mathring{v_j}))
\end{displaymath}}where $\gamma$ is a margin separating the positive and negative triplets. The model parameters are learned by optimizing the above loss using stochastic gradient descent with a mini-batch based on the Adam optimizer.


\subsection{Training Regime}
\label{sec:regime}
Given $\widetilde{G_{\text{tr}}}=(\sV_{\text{tr}},\sR_{\text{tr}},\sE_{\text{tr}})$, we divide $\sE_{\text{tr}}$ into $\sF_{\text{tr}}$ and $\sT_{\text{tr}}$ with a ratio of 3:1. For every epoch, we randomly re-split $\sF_{\text{tr}}$ and $\sT_{\text{tr}}$ with the minimal constraint that $\sF_{\text{tr}}$ includes the minimum spanning tree of $\widetilde{G_{\text{tr}}}$ and $\sF_{\text{tr}}$ covers all relations in $\sR_{\text{tr}}$ so that all entity and relation embedding vectors are appropriately learned. At the beginning of each epoch, we initialize all feature vectors using Glorot initialization. 

This dynamic split and re-initialization strategy allow \ours to robustly learn the model parameters, which makes the model more easily generalizable to an inference graph. In Section~\ref{sec:abl}, we empirically observe the importance of dynamic split by the ablation study of \ours. Since we randomly re-initialize all feature vectors per epoch during training, \ours learns how to compute embedding vectors using random feature vectors, and this is beneficial for computing embeddings with random features at inference time. This observation is consistent with recent studies in~\cite{rni,rgin} showing that the expressive power of GNNs can be enhanced with randomized initial node features. However,~\cite{rni,rgin} have analyzed GNNs for standard graphs but not for knowledge graphs. We will further investigate the effects of the combination of dynamic split and random re-initialization strategy from a theoretical point of view.


\subsection{Embedding of New Relations and Entities}
\label{sec:new}
Given $G_{\text{inf}}=(\sV_{\text{inf}},\sR_{\text{inf}},\sF_{\text{inf}})$, we create the relation graph discussed in Section~\ref{sec:rgraph} and compute the relation and entity embedding vectors using the learned model parameters of \ours. Algorithm~\ref{alg:emb} shows the overall procedure. Based on the generated embedding vectors of entities and relations on $G_{\text{inf}}$, we can predict missing triplets. For example, to solve ($v_i$, $r_k$, ?), we plug each entity $v_j\in\sV_{\text{inf}}$ into the given triplet and compute the score using (\ref{eq:score}) where $\mtW$ is already trained during training. The missing entity is predicted to be the one with the highest score.

\begin{algorithm}[t]
\small
\caption{Embeddings via \ours at Inference Time}
\label{alg:emb}
\begin{flushleft}
\textbf{Input}: $G_{\text{inf}}=(\sV_{\text{inf}},\sR_{\text{inf}},\sF_{\text{inf}})$, the trained model parameters: $\mH$, $\mW^{(l)}$, $\mTh^{(l)}$, $\vy^{(l)}$, $c_1^{(l)},{\cdots}, c_B^{(l)}$ for $l=0,{\cdots},L-1$, $\mhH$, $\mhW^{(\widehat{l})}$, $\mhTh^{(\widehat{l})}$, $\vhy^{(\widehat{l})}$, $\widehat{l}=0,{\cdots},\widehat{L}-1$, $\mM$, $\mhM$\\
\textbf{Output}: $\vz_i$ for all $r_i\in\sR_{\text{inf}}$ and $\vh_j$ for all $v_j\in\sV_{\text{inf}}$
\end{flushleft}
\begin{algorithmic}[1] 
\STATE Create the relation graph $\mA$ as discussed in Section~\ref{sec:rgraph}.
\STATE Initialize $\vx_i$ for all $r_i\in\sR_{\text{inf}}$ and $\vhx_j$ for all $v_j\in\sV_{\text{inf}}$ using Glorot initialization.
\STATE Set $\vz_i^{(0)}\leftarrow \mH\vx_i$ and $\vh_j^{(0)}\leftarrow \mhH\vhx_j$.
\FOR{$l=0,{\cdots},L-1$}
	\FOR{$r_i\in\sR_{\text{inf}}$}
	\STATE Compute $\vz_i^{(l+1)}$ according to (\ref{eq:relup}) using $\mW^{(l)}$, $\mTh^{(l)}$, $\vy^{(l)}$, and $c_1^{(l)},{\cdots},c_B^{(l)}$.
	\ENDFOR
\ENDFOR
\FOR{$l=0,{\cdots},\widehat{L}-1$}
	\FOR{$v_j\in\sV_{\text{inf}}$}
	\STATE Compute $\vh_j^{(l+1)}$ according to (\ref{eq:nodeup}) using $\mhW^{(l)}$, $\mhTh^{(l)}$, $\vhy^{(l)}$, and $\vz_i^{(L)}$ for $r_i\in\sR_{\text{inf}}$.
	\ENDFOR
\ENDFOR
\STATE $\vz_i\leftarrow \mM \vz_i^{(L)}$ for $r_i\in\sR_{\text{inf}}$ and $\vh_j\leftarrow \mhM\vh_j^{(\widehat{L})}$ for $v_j\in\sV_{\text{inf}}$.
\end{algorithmic}
\end{algorithm}

\section{Experimental Results}
\label{sec:exp}
We compare the performance of \ours with other inductive knowledge graph completion methods.

\begin{table*}[t]
\scriptsize
\centering
\caption{Inductive link prediction performance on 12 different datasets, where all entities are new, whereas the last digits of each dataset (100, 75, 50, and 25) indicate the ratio of new relations. The best results are boldfaced and the second-best results are underlined. Our model, \ours, significantly outperforms the baseline methods in most cases.}
\label{tb:main}
\setlength{\tabcolsep}{0.61em}
\begin{tabular}{ccccccccccccccccccc}
\toprule
 & \multicolumn{4}{c}{\nld} & \multicolumn{4}{c}{\nlc} & \multicolumn{4}{c}{\nlb} & \multicolumn{4}{c}{\nla} \\
 & MR & MRR & Hit@10 & Hit@1 & MR & MRR & Hit@10 & Hit@1 & MR & MRR & Hit@10 & Hit@1 & MR & MRR & Hit@10 & Hit@1 \\
\midrule
\grail & 928.4 & 0.135 & 0.173 & 0.114 & 526.0 & 0.096 & 0.205 & 0.036 & 837.6 & 0.162 & 0.288 & 0.104 & 692.9 & 0.216 & 0.366 & 0.160 \\
\compile & 743.1 & 0.123 & 0.209 & 0.071 & 519.6 & 0.178 & \underline{0.361} & 0.093 & 466.6 & 0.194 & 0.330 & 0.125 & 438.9 & 0.189 & 0.324 & 0.115 \\
\snri & 809.8 & 0.042 & 0.064 & 0.029 & 418.7 & 0.088 & 0.177 & 0.040 & 584.6 & 0.130 & 0.187 & 0.095 & 417.7 & 0.190 & 0.270 & 0.140 \\
\indigo & 621.1 & 0.160 & 0.247 & 0.109 & 587.4 & 0.121 & 0.156 & 0.098 & 864.9 & 0.167 & 0.217 & 0.134 & 812.4 & 0.166 & 0.206 & 0.134 \\
\rmpi & \underline{143.9} & \underline{0.220} & 0.376 & \underline{0.136} & 244.5 & 0.138 & 0.275 & 0.061 & 479.1 & 0.185 & 0.307 & 0.109 & 385.7 & 0.213 & 0.329 & 0.130 \\
\cgcn & 877.9 & 0.008 & 0.014 & 0.001 & 750.5 & 0.014 & 0.025 & 0.003 & 1183.6 & 0.003 & 0.005 & 0.000 & 1052.5 & 0.006 & 0.010 & 0.000 \\
\npiece & 755.1 & 0.012 & 0.018 & 0.004 & 565.8 & 0.042 & 0.081 & 0.020 & 832.2 & 0.037 & 0.079 & 0.013 & 620.9 & 0.098 & 0.166 & 0.057\\
\nelp & 530.3 & 0.084 & 0.181 & 0.035 & 447.3 & 0.117 & 0.273 & 0.048 & 802.4 & 0.101 & 0.190 & 0.064 & 631.8 & 0.148 & 0.271 & 0.101 \\
\drum & 532.6 & 0.076 & 0.138 & 0.044 & 445.4 & 0.152 & 0.313 & 0.072 & 803.8 & 0.107 & 0.193 & 0.070 & 637.1 & 0.161 & 0.264 & 0.119 \\
\blp & 564.8 & 0.019 & 0.037 & 0.006 & \underline{242.5} & 0.051 & 0.120 & 0.012 & 426.5 & 0.041 & 0.093 & 0.011 & 332.9 & 0.049 & 0.095 & 0.024 \\
\qblp & 754.6 & 0.004 & 0.003 & 0.000 & 258.8 & 0.040 & 0.095 & 0.007 & 383.6 & 0.048 & 0.097 & 0.020 & \underline{287.2} & 0.073 & 0.151 & 0.027 \\
\nbf & 208.2 & 0.096 & 0.199 & 0.032 & 256.2 & 0.137 & 0.255 & 0.077 & \underline{332.0} & \underline{0.225} & \underline{0.346} & \underline{0.161} & 421.8 & \underline{0.283} & \underline{0.417} & \underline{0.224} \\
\red & 201.7 & 0.212 & \underline{0.385} & 0.114 & 470.1 & \underline{0.203} & 0.353 & \underline{0.129} & 622.5 & 0.179 & 0.280 & 0.115 & 403.0 & 0.214 & 0.266 & 0.166 \\
\raild & 598.1 & 0.018 & 0.037 & 0.005 & N/A & N/A & N/A & N/A & N/A & N/A & N/A & N/A & N/A & N/A & N/A & N/A \\
\ours & \textbf{92.6} & \textbf{0.309} & \textbf{0.506} & \textbf{0.212} & \textbf{59.1} & \textbf{0.261} & \textbf{0.464} & \textbf{0.167} & \textbf{105.1} & \textbf{0.281} & \textbf{0.453} & \textbf{0.193} & \textbf{90.1} & \textbf{0.334} & \textbf{0.501} & \textbf{0.241} \\
\midrule
 & \multicolumn{4}{c}{\wkd} & \multicolumn{4}{c}{\wkc} & \multicolumn{4}{c}{\wkb} & \multicolumn{4}{c}{\wka} \\
 & MR & MRR & Hit@10 & Hit@1 & MR & MRR & Hit@10 & Hit@1 & MR & MRR & Hit@10 & Hit@1 & MR & MRR & Hit@10 & Hit@1 \\
\midrule
\cgcn & 5861.5 & 0.003 & 0.009 & 0.000 & 1265.1 & 0.015 & 0.028 & 0.003 & 3297.4 & 0.003 & 0.002 & 0.001 & 1591.7 & 0.009 & 0.020 & 0.000 \\
\npiece & 5334.1 & 0.007 & 0.018 & 0.002 & 800.3 & 0.021 & 0.052 & 0.003 & 3256.4 & 0.008 & 0.013 & 0.002 & 814.5 & 0.053 & 0.122 & 0.019 \\
\nelp & 5665.5 & 0.009 & 0.016 & 0.005 & 1191.5 & 0.020 & 0.054 & 0.004 & 4160.8 & 0.025 & 0.054 & 0.007 & 1384.1 & 0.068 & 0.104 & 0.046 \\
\drum & 5668.0 & 0.010 & 0.019 & 0.004 & 1192.1 & 0.020 & 0.043 & 0.007 & 4163.0 & 0.017 & 0.046 & 0.002 & 1383.2 & 0.064 & 0.116 & 0.035 \\
\blp & 3888.1 & 0.012 & 0.025 & 0.003 & \underline{523.9} & 0.043 & 0.089 & 0.016 & 1625.7 & 0.041 & 0.092 & 0.013 & \textbf{175.4} & 0.125 & 0.283 & 0.055 \\
\qblp & 2863.1 & 0.012 & 0.025 & 0.003 & 555.0 & 0.044 & 0.091 & 0.016 & \textbf{1371.4} & 0.035 & 0.080 & 0.011 & 342.0 & 0.116 & 0.294 & 0.042 \\
\nbf & 4030.3 & 0.014 & 0.026 & 0.005 & 548.1 & 0.072 & 0.172 & 0.028 & 2874.0 & \underline{0.062} & \underline{0.105} & \textbf{0.036} & 790.5 & 0.154 & \underline{0.301} & 0.092 \\
\red & 5382.4 & \underline{0.096} & \underline{0.136} & \underline{0.070} & 906.2 & \underline{0.172} & \underline{0.290} & \underline{0.110} & 3198.3 & 0.058 & 0.093 & 0.033 & 769.2 & \underline{0.170} & 0.263 & \underline{0.111} \\
\raild & \underline{2005.6} & 0.026 & 0.052 & 0.010 & N/A & N/A & N/A & N/A & N/A & N/A & N/A & N/A & N/A & N/A & N/A & N/A \\
\ours & \textbf{1515.7} & \textbf{0.107} & \textbf{0.169} & \textbf{0.072} & \textbf{315.5} & \textbf{0.247} & \textbf{0.362} & \textbf{0.179} & \underline{1374.1} & \textbf{0.068} & \textbf{0.135} & \underline{0.034} & \underline{263.8} & \textbf{0.186} & \textbf{0.309} & \textbf{0.124} \\
\midrule
 & \multicolumn{4}{c}{\fbd} & \multicolumn{4}{c}{\fbc} & \multicolumn{4}{c}{\fbb} & \multicolumn{4}{c}{\fba} \\
 & MR & MRR & Hit@10 & Hit@1 & MR & MRR & Hit@10 & Hit@1 & MR & MRR & Hit@10 & Hit@1 & MR & MRR & Hit@10 & Hit@1 \\
\midrule
\cgcn & 1201.2 & 0.015 & 0.025 & 0.008 & 1211.6 & 0.013 & 0.026 & 0.000 & 2193.1 & 0.004 & 0.006 & 0.002 & 1957.4 & 0.003 & 0.004 & 0.000 \\
\npiece & 1131.3 & 0.006 & 0.009 & 0.001 & 1162.7 & 0.016 & 0.029 & 0.007 & 1314.3 & 0.021 & 0.048 & 0.006 & 916.3 & 0.044 & 0.114 & 0.011 \\
\nelp & 988.2 & 0.026 & 0.057 & 0.007 & 855.0 & 0.056 & 0.099 & 0.030 & 1501.9 & 0.088 & 0.184 & 0.043 & 997.8 & 0.164 & 0.309 & 0.098 \\
\drum & 984.0 & 0.034 & 0.077 & 0.011 & 853.8 & 0.065 & 0.121 & 0.034 & 1490.2 & 0.101 & 0.191 & 0.061 & 992.5 & \underline{0.175} & \underline{0.320} & \underline{0.109} \\
\blp & 913.1 & 0.017 & 0.035 & 0.004 & \underline{705.1} & 0.047 & 0.085 & 0.024 & 588.5 & 0.078 & 0.156 & 0.037 & 384.5 & 0.107 & 0.212 & 0.053 \\
\qblp & 842.8 & 0.013 & 0.026 & 0.003 & 798.3 & 0.041 & 0.084 & 0.017 & \textbf{564.9} & 0.071 & 0.147 & 0.030 & \underline{352.6} & 0.104 & 0.226 & 0.043 \\
\nbf & 451.5 & 0.072 & 0.154 & 0.026 & 550.8 & 0.089 & 0.166 & 0.048 & 758.5 & \textbf{0.130} & \textbf{0.259} & \underline{0.071} & 571.4 & \textbf{0.224} & \textbf{0.410} & \textbf{0.137} \\
\red & \underline{375.6} & \underline{0.121} & \underline{0.263} & \underline{0.053} & 890.0 & \underline{0.107} & \underline{0.201} & \underline{0.057} & 1169.3 & \underline{0.129} & \underline{0.251} & \textbf{0.072} & 1234.1 & 0.145 & 0.284 & 0.077 \\
\raild & 686.0 & 0.031 & 0.048 & 0.016 & N/A & N/A & N/A & N/A & N/A & N/A & N/A & N/A & N/A & N/A & N/A & N/A \\
\ours & \textbf{171.5} & \textbf{0.223} & \textbf{0.371} & \textbf{0.146} & \textbf{217.4} & \textbf{0.189} & \textbf{0.325} & \textbf{0.119} & \underline{580.2} & 0.117 & 0.218 & 0.067 & \textbf{330.3} & 0.133 & 0.271 & 0.067 \\
\bottomrule
\end{tabular}
\end{table*}

\subsection{Datasets and Baseline Methods}
\label{sec:data}
Since existing datasets do not contain new relations at $\widetilde{G_{\text{inf}}}$, we create 12 datasets using three benchmarks, \nell~\cite{nell995},\wk~\cite{raild}, and \fb~\cite{fb}. Let us call these benchmarks NL, WK, and FB, respectively. For each benchmark, we create four datasets by varying the percentage of triplets with new relations: 100\%, 75\%, 50\% and 25\%. For example, in \nlc, approximately 75\% of triplets have new relations, and 25\% of triplets have known relations, i.e., semi-inductive inference for relations. On the other hand, in \nld, all triplets have new relations, i.e., inductive inference for relations. In all 12 datasets, all entities in $\widetilde{G_{\text{inf}}}$ are new entities, as also assumed in~\cite{grail}. Details about these datasets are described in Appendix~\ref{app:data}.

We compare the performance of \ours with 14 different methods: \grail~\cite{grail}, \compile~\cite{compile}, \snri~\cite{snri}, \indigo~\cite{indigo}, \rmpi~\cite{rmpi}, \cgcn~\cite{cgcn}, \npiece~\cite{npiece}, \nelp~\cite{nelp}, \drum~\cite{drum}, \blp~\cite{blp}, \qblp~\cite{qblp}, \nbf~\cite{nbf}, \red~\cite{red}, and \raild~\cite{raild}. 

Since the original implementations of \grail, \compile, \snri, \indigo, and \rmpi were based on subgraph sampling, we extended them to consider all entities in $\sV_{\text{inf}}$ to evaluate the performance more accurately. Due to the scalability issues of \grail, \compile, \snri, \indigo, and \rmpi, they only run on NL datasets. \blp, \qblp and \raild require BERT-based pre-trained vectors which are produced based on the names and textual descriptions of entities or relations. We provide \blp, \qblp and \raild with the pre-trained vectors using available information. Since \raild is not implemented for the case where both known and new relations are present at inference time, we could not report the results of \raild for the 75\%, 50\% and 25\% settings. We provide the same training graph for all methods. Given $\sE_{\text{tr}}$, how to use $\sE_{\text{tr}}$ depends on each method. For example, \grail uses the entire $\sE_{\text{tr}}$ for training without any split. In \ours, we use the dynamic split scheme as described in Section~\ref{sec:regime}. For a fair comparison, we feed $\sF_{\text{inf}}$, $\sT_{\text{val}}$ and $\sT_{\text{test}}$ to all baselines exactly in the same way \ours uses. More details about how we run the baseline methods are described in Appendix~\ref{app:base}. 

We set $d=32$ and $\widehat{d}=32$ for \ours and all the baseline methods. Note that we initialize the initial features of relations and entities ($\vx_i$ in Section~\ref{sec:rel} and $\vhx_i$ in Section~\ref{sec:node}) using Glorot initialization in \ours. Details about how we tune the hyperparameters of \ours are described in Appendix~\ref{app:hyper}.

\begin{table}[t]
\scriptsize
\centering
\caption{Inductive link prediction performance of known relations and new relations on \nlb.}
\label{tb:kn}
\begin{tabular}{ccccccc}
\toprule
 & \multicolumn{3}{c}{Known Relations} & \multicolumn{3}{c}{New Relations} \\
 & MR & MRR & Hit@10 & MR & MRR & Hit@10 \\
\midrule
\grail & 711.8 & \underline{0.264} & \underline{0.389} & 936.0 & 0.082 & 0.209 \\
\compile & 418.0 & 0.250 & 0.383 & 504.6 & 0.150 & 0.288 \\
\snri & 515.8 & 0.206 & 0.240 & 638.4 & 0.071 & 0.146 \\
\indigo & 943.6 & 0.185 & 0.264 & 803.4 & 0.152 & 0.180 \\
\rmpi & 492.8 & 0.192 & 0.312 & 468.4 & 0.180 & 0.304 \\
\cgcn & 1195.6 & 0.004 & 0.005 & 1174.2 & 0.003 & 0.004 \\
\npiece & 575.6  & 0.067  & 0.151  & 1033.5  & 0.016  & 0.039 \\
\nelp & 784.5 & 0.147 & 0.204 & 816.3 & 0.065 & 0.180 \\
\drum & 787.8 & 0.146 & 0.196 & 816.4 & 0.076 & 0.191 \\
\blp & 357.1 & 0.056 & 0.131 & 480.8 & 0.030 & 0.062 \\
\qblp & \underline{271.2} & 0.073 & 0.142 & 471.4 & 0.029 & 0.061 \\
\nbf & 317.4  & 0.231  & 0.353  & \underline{350.7}  & \underline{0.217}  & \underline{0.338} \\
\red & 565.3 & 0.210 & 0.300 & 667.2 & 0.154 & 0.265 \\
\ours & \textbf{100.7} & \textbf{0.330} & \textbf{0.481} & \textbf{108.5} & \textbf{0.244} & \textbf{0.430} \\
\bottomrule
\end{tabular}
\end{table}

\begin{table}[t]
\scriptsize
\centering
\caption{Inductive link prediction on \nlz and \nlg, where all relations are known and all entities are new.}
\label{tb:trans}
\begin{tabular}{cccccc}
\toprule
& & MR & MRR & Hit@10 & Hit@1 \\
\midrule
\multirow{15}{*}{\nlz} & \grail & 508.2 & 0.192 & 0.332 & 0.114 \\
& \compile & 561.4 & 0.229 & 0.381 & 0.147 \\
& \snri & 561.3 & 0.117 & 0.176 & 0.081 \\
& \indigo & 705.6 & 0.201 & 0.263 & 0.166 \\
& \rmpi & 396.1  & 0.225  & 0.339  & 0.158 \\
& \cgcn & 954.3 & 0.005 & 0.009 & 0.001 \\
& \npiece & 345.2 & 0.094 & 0.210 & 0.037 \\
& \nelp & 566.3 & 0.175 & 0.326 & 0.102 \\
& \drum & 565.9 & 0.200 & 0.343 & 0.130 \\
& \blp & 467.2 & 0.044 & 0.100 & 0.011 \\
& \qblp & 346.2 & 0.060 & 0.144 & 0.013 \\
& \nbf & \underline{160.2} & \underline{0.263} & \underline{0.430} & \underline{0.177} \\
& \red & 330.9 & 0.222 & 0.368 & 0.147 \\
& \raild & 468.4 & 0.050 & 0.109 & 0.014 \\
& \ours & \textbf{152.4} & \textbf{0.269} & \textbf{0.431} & \textbf{0.189} \\
\midrule
\multirow{15}{*}{\nlg} & \grail & 18.7  & 0.499  & 0.595  & 0.405 \\
& \compile & 20.1  & 0.474  & 0.575  & 0.390 \\
& \snri & 21.2  & 0.419  & 0.520  & 0.330 \\
& \indigo & 20.4  & 0.521  & 0.595  & 0.495 \\
& \rmpi & 50.3  & 0.484  & 0.545  & 0.425 \\
& \cgcn & 11.8  & 0.282  & 0.750  & 0.005 \\
& \npiece & 9.8  & \underline{0.677}  & \underline{0.885}  & \underline{0.550} \\
& \nelp & 33.0  & 0.547  & 0.785  & 0.400 \\
& \drum & 33.4  & 0.536  & 0.760  & 0.400 \\
& \blp & 42.0  & 0.169  & 0.470  & 0.055 \\
& \qblp & 18.8  & 0.326  & 0.545  & 0.230 \\
& \nbf & \underline{7.1}  & 0.613  & 0.875  & 0.500 \\
& \red & 15.0  & 0.544  & 0.705  & 0.470 \\
& \raild & 113.6  & 0.052  & 0.205  & 0.000 \\
& \ours & \textbf{6.0}  & \textbf{0.739}  & \textbf{0.895}  & \textbf{0.660} \\
\bottomrule
\end{tabular}
\end{table}

\subsection{Inductive Link Prediction}
\label{sec:main_exp}
We measure the inductive link prediction performance of the methods using standard metrics~\cite{kgs}: MR ($\downarrow$), MRR ($\uparrow$), Hit@10 ($\uparrow$), and Hit@1 ($\uparrow$). Table~\ref{tb:main} shows the results on 12 different datasets, where all entities are new, and each dataset has a different ratio of new relations. Among the 12 datasets, \nld, \wkd, and \fbd have entirely new relations (i.e., inductive inference for relations), whereas the other 9 datasets contain a mixture of new and known relations (i.e., semi-inductive inference for relations). In Table~\ref{tb:main}, we first note that \ours significantly outperforms all the baseline methods in \nld, \wkd, and \fbd, which are the most challenging datasets since relations and entities are all new. In these datasets, the performance gap between \ours and the best baseline method is considerable in terms of all metrics. This shows that \ours is the most effective method for inductive inference for relations.

Let us now consider the semi-inductive inference settings. In all NL and WK datasets as well as \fbc datasets, \ours significantly outperforms the baseline methods. We also note that the performance gap between \ours and the baseline methods becomes more prominent when the ratio of new relations increases.

While \ours shows clearly better performance than the baseline methods on 10 out of 12 datasets, some baseline methods such as \nbf and \red show better performance than \ours on \fba and \fbb. Indeed, we notice that there exist simple rules between known relations in these datasets, and thus, even a simple rule-based prediction works well on these datasets. This partly explains the performance of \nelp, \drum, \red, and \nbf, which are designed to capture rules or patterns between known relations and directly apply them at inference time. Different from these methods, \ours does not memorize particular patterns between known relations; instead, \ours focuses more on generalizability which is more beneficial for generating embeddings of new relations.

In Table~\ref{tb:kn}, we analyze the model performances on triplets with known relations and new relations on \nlb. All methods perform better on known relations than new relations. Also, for the baseline methods, the performance gaps between known and new relations are substantial. We see that the performance of \ours is much better than the best baseline methods in all metrics for both known and new relations.

\subsection{Inductive Link Prediction with Known Relations}
Even though \ours is designed to consider the case where new relations appear at inference time, we also conduct experiments on the conventional inductive link prediction scenario where all relations are known and all entities are new~\cite{grail}. We create a dataset \nlz that satisfies this constraint, where $|\sV_\text{tr}| = 1,814$, $|\sR_\text{tr}| = 134$, $|\sE_\text{tr}| = 7,796$ and $|\sV_\text{inf}| = 2,026$, $|\sR_\text{inf}| = 112$, $|\sE_\text{inf}| = 3,813$. We call this dataset \nlz since it includes 0\% new relations.

Also, we take an existing benchmark dataset, \nlg, from~\cite{grail}. We note that different papers did experiments under different settings, even though they used the same benchmark dataset. We conduct our experiments in a setting consistent with our other experiments. For the results on \nlg, our reproduced results can differ from those reported in the previous literature for the following reasons: (i) we use the validation set inside the ``ind-test'' set provided in \nlg, (ii) when measuring the link prediction performance, we use the test set of ``ind-test'', (iii) for a fair comparison, we set the embedding dimension to 32 for all methods, and (iv) when conducting link prediction, we set the candidate set to be the entire entity set of the inference graph.

Table~\ref{tb:trans} shows the inductive link prediction results on \nlz and \nlg, where all entities are new and all relations are known. We see that \ours outperforms all baselines in all metrics. Even though \ours does not learn relation-specific patterns as other baselines do, \ours shows reasonable performance on transductive inference for relations while having extra generalization capability to semi-inductive and inductive inferences for relations.

\begin{table}[t]
\scriptsize
\centering
\caption{Ablation Studies of \ours.}
\label{tb:abl}
\setlength{\tabcolsep}{0.64em}
\begin{tabular}{ccccccc}
\toprule
 & \multicolumn{2}{c}{\nld} & \multicolumn{2}{c}{\wkd} & \multicolumn{2}{c}{\fbd} \\
 & MRR & Hit@10 & MRR & Hit@10 & MRR & Hit@10 \\
\midrule
w/ mean aggregator & 0.259 & 0.421 & 0.047 & 0.106 & 0.110 & 0.184 \\
w/ sum aggregator & 0.049 & 0.082 & 0.000 & 0.000 & 0.001 & 0.000 \\
w/ self-loop vector & 0.133 & 0.292 & 0.007 & 0.015 & 0.014 & 0.027 \\
w/o dynamic split   & 0.234 & 0.418 & 0.036 & 0.100 & 0.134 & 0.271 \\
w/o relation update & 0.235 & 0.415 & 0.057 & 0.138 & 0.183 & 0.310 \\
w/o binning & 0.209 & 0.443 & 0.070 & 0.138 & 0.142 & 0.292 \\
\ours & 0.309 & 0.506 & 0.107 & 0.169 & 0.223 & 0.371 \\
\bottomrule
\end{tabular}
\end{table}

\begin{figure*}[t]
\centering
\begin{subfigure}{0.33\textwidth}
  \centering
  \includegraphics[height=3.8cm]{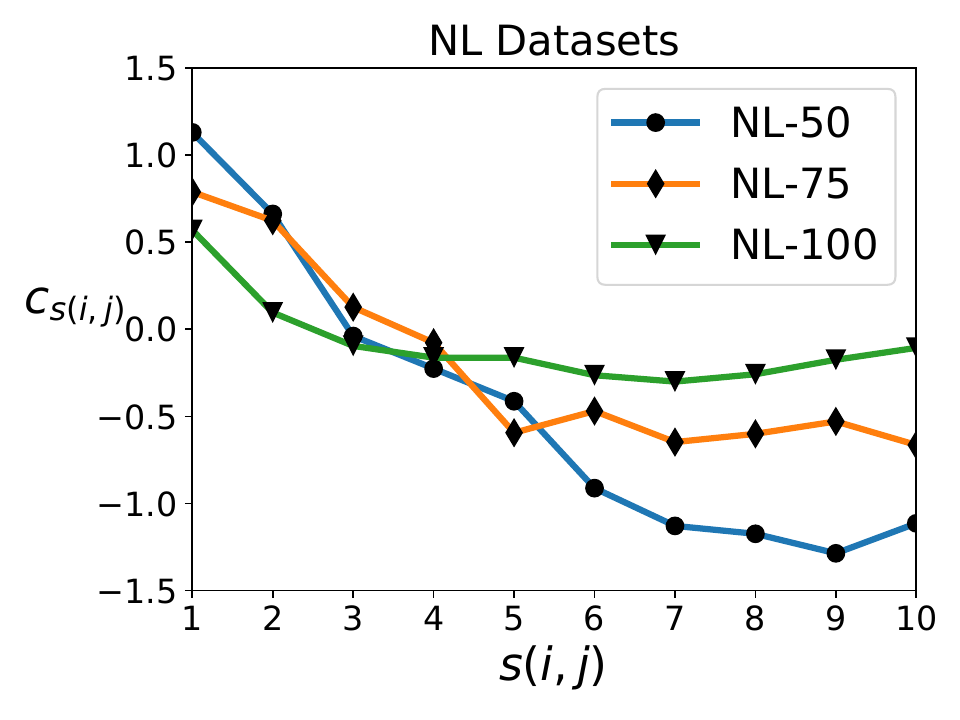}
  \caption{Results on NL Datasets}
\end{subfigure}
\begin{subfigure}{0.33\textwidth}
  \centering
  \includegraphics[height=3.8cm]{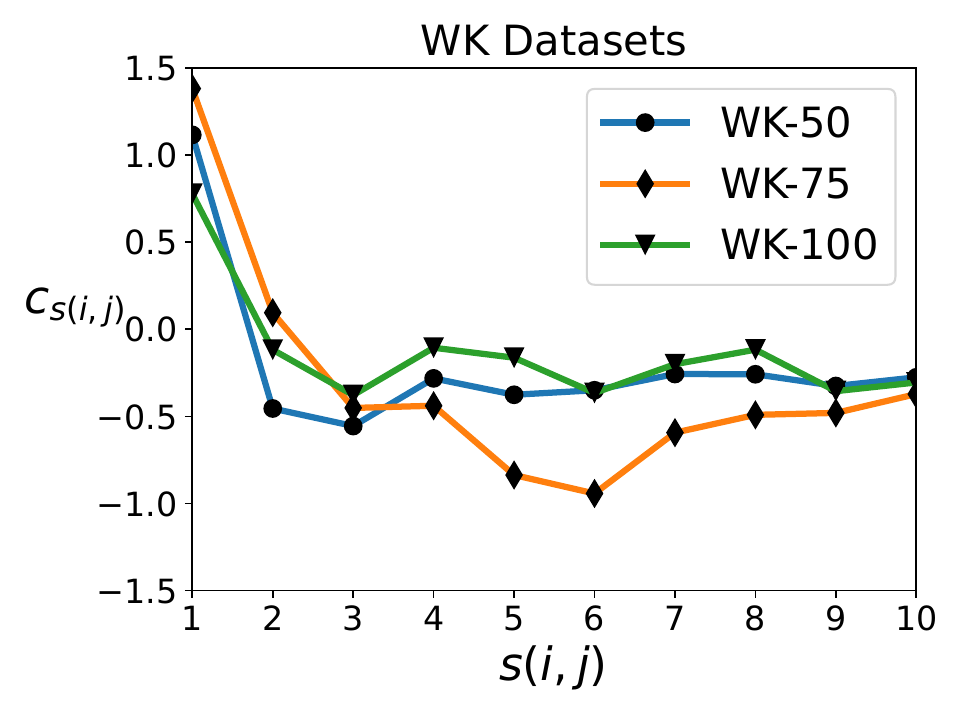}
  \caption{Results on WK Datasets}
\end{subfigure}%
\begin{subfigure}{0.33\textwidth}
  \centering
  \includegraphics[height=3.8cm]{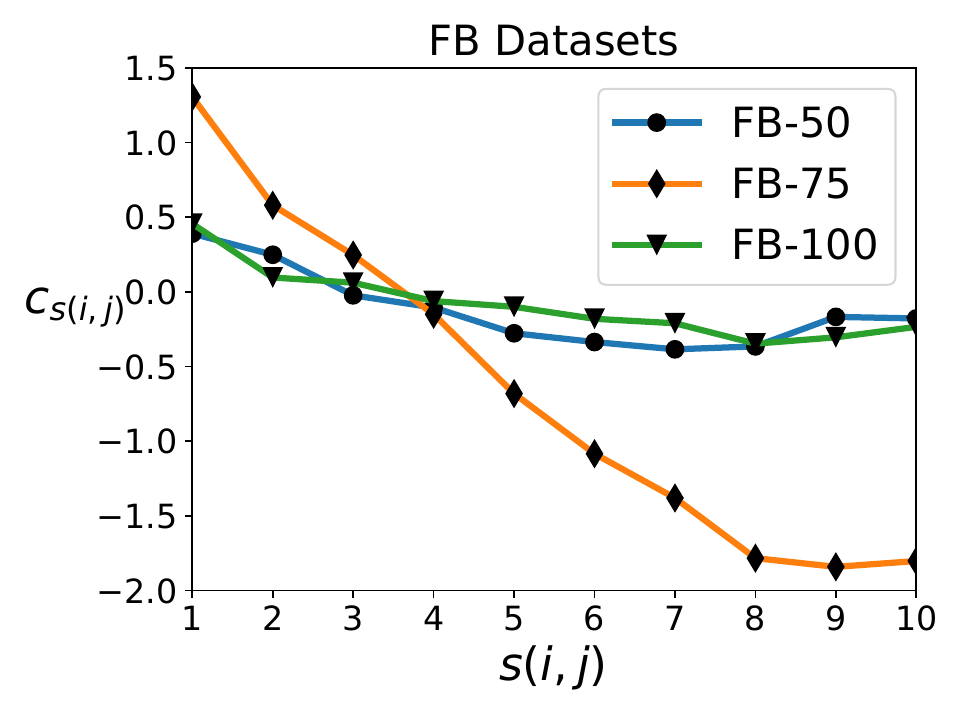}
  \caption{Results on FB Datasets}
\end{subfigure}
\caption{Visualization of the $c_{s(i,j)}$ values learned by \ours with $B=10$. The $c_{s(i,j)}$ value is expected to be large for a small $s(i,j)$, and small for a large $s(i,j)$. See Section~\ref{sec:rel} for more details.}
\label{fig:bin}
\end{figure*}

\subsection{Ablation Studies of \ours}
\label{sec:abl}
We conduct ablation studies for \ours to validate the importance of each module of \ours. Specifically, we compare the performance of \ours under the following settings: (i) we replace the attention-based aggregation in (\ref{eq:relup}) and (\ref{eq:nodeup}) with the mean aggregator or (ii) the sum aggregator; (iii) when we update an entity representation in (\ref{eq:nodeup}), we replace $\bar{\vz}_i^{(L)}$ with a separate learnable self-loop vector as in~\cite{cgcn}; (iv) we do not dynamically re-split $\sF_{\text{tr}}$ and $\sT_{\text{tr}}$ as described in Section~\ref{sec:regime}, i.e., we fix the split during training; (v) we do not update a relation representation vector; (vi) we set $B=1$ in (\ref{eq:indx}), i.e., we do not utilize the affinity weights in updating relation representations. The results of these settings are shown in Table~\ref{tb:abl} in order. Removing each module leads to a noticeable degradation in the performance of \ours. While some are more critical and others are less, \ours achieves the best performance when all modules come together.

\subsection{Qualitative Analysis of \ours}
\label{sec:qual}
In Section~\ref{sec:rel}, when updating the relation vectors using (\ref{eq:relatt}), we introduce the learnable parameters $c_{s(i,j)}$ for computing the attention coefficient between $r_i$ and $r_j$. By definition of $s(i,j)$ in (\ref{eq:indx}), the value of $s(i,j)$ is small if $r_i$ and $r_j$ are similar. Since we expect that two similar relations have a high attention value, we expect that $c_{s(i,j)}$ is large for a small $s(i,j)$, and $c_{s(i,j)}$ is small for a large $s(i,j)$. Figure~\ref{fig:bin} shows the learned $c_{s(i,j)}$ values according to $s(i,j)$. Even though some exceptions exist, overall, the plots are going down from left to right; the $c_{s(i,j)}$ values are learned as expected. Indeed, when we tune the number of bins $B\in \{1, 5, 10\}$, \ours achieves the best performance when $B=10$, showing the advantage of differentiating bins.

\section{Conclusion and Future Work}
We consider challenging and realistic inductive learning scenarios where new entities accompany new relations. Our method, \ours, can generate embeddings of new relations and entities only appearing at inference time. \ours conducts inferences based only on the structure of a given knowledge graph without any extra information about the entities and relations or the aid of rich language models. 

While existing methods are biased toward learning the patterns of known relations, \ours focuses more on the generalization capability useful for modeling new relations. We will investigate the ways in which \ours can incorporate known-relation-specific patterns into inferences when known relations are dominant. We also plan to do the theoretical analysis of \ours as done in~\cite{gin},~\cite{log} and~\cite{gsage}, as well as consider some extensions to hyper-relational facts~\cite{hynt} and bi-level or hierarchical structures~\cite{bive,robokg}. Finally, we will look into how we can make the predictions of \ours robust and reliable to possibly noisy information~\citep{acl} in a given knowledge graph.

\section*{Acknowledgements}
This research was supported by an NRF grant funded by MSIT 2022R1A2C4001594 (Extendable Graph Representation Learning) and an IITP grant funded by MSIT 2022-0-00369 (Development of AI Technology to support Expert Decision-making that can Explain the Reasons/Grounds for Judgment Results based on Expert Knowledge). 
\bibliography{ingram_icml2023_bib_camera}

\begin{thebibliography}{53}
\providecommand{\natexlab}[1]{#1}
\providecommand{\url}[1]{\texttt{#1}}
\expandafter\ifx\csname urlstyle\endcsname\relax
  \providecommand{\doi}[1]{doi: #1}\else
  \providecommand{\doi}{doi: \begingroup \urlstyle{rm}\Url}\fi

\bibitem[Abboud et~al.(2021)Abboud, Ceylan, Grohe, and Lukasiewicz]{rni}
Abboud, R., Ceylan, {\.I}.~{\.I}., Grohe, M., and Lukasiewicz, T.
\newblock The surprising power of graph neural networks with random node
  initialization.
\newblock In \emph{Proceedings of the 30th International Joint Conference on
  Artificial Intelligence}, pp.\  2112--2118, 2021.

\bibitem[Ali et~al.(2021)Ali, Berrendorf, Galkin, Thost, Ma, Tresp, and
  Lehmann]{qblp}
Ali, M., Berrendorf, M., Galkin, M., Thost, V., Ma, T., Tresp, V., and Lehmann,
  J.
\newblock Improving inductive link prediction using hyper-relational facts.
\newblock In \emph{Proceedings of the 20th International Semantic Web
  Conference}, pp.\  74--92, 2021.

\bibitem[Barcelo et~al.(2022)Barcelo, Galkin, Morris, and Orth]{log}
Barcelo, P., Galkin, M., Morris, C., and Orth, M.~R.
\newblock Weisfeiler and {L}eman go relational.
\newblock In \emph{Proceedings of the 1st Learning on Graphs Conference}, 2022.

\bibitem[Brody et~al.(2022)Brody, Alon, and Yahav]{gat2}
Brody, S., Alon, U., and Yahav, E.
\newblock How attentive are graph attention networks?
\newblock In \emph{Proceedings of the 10th International Conference on Learning
  Representations}, 2022.

\bibitem[Chen et~al.(2021)Chen, He, Wu, and Wang]{tact}
Chen, J., He, H., Wu, F., and Wang, J.
\newblock Topology-aware correlations between relations for inductive link
  prediction in knowledge graphs.
\newblock In \emph{Proceedings of the 35th AAAI Conference on Artificial
  Intelligence}, pp.\  6271--6278, 2021.

\bibitem[Chung \& Whang(2023)Chung and Whang]{bive}
Chung, C. and Whang, J.~J.
\newblock Learning representations of bi-level knowledge graphs for reasoning
  beyond link prediction.
\newblock In \emph{Proceedings of the 37th AAAI Conference on Artificial
  Intelligence}, pp.\  4208--4216, 2023.

\bibitem[Chung et~al.(2023)Chung, Lee, and Whang]{hynt}
Chung, C., Lee, J., and Whang, J.~J.
\newblock Representation learning on hyper-relational and numeric knowledge
  graphs with transformers.
\newblock In \emph{Proceedings of the 29th ACM SIGKDD Conference on Knowledge
  Discovery and Data Mining}, pp.\  310--322, 2023.

\bibitem[Cui et~al.(2022)Cui, Wang, Sun, Liu, Jiang, Han, and Hu]{argcn}
Cui, Y., Wang, Y., Sun, Z., Liu, W., Jiang, Y., Han, K., and Hu, W.
\newblock Inductive knowledge graph reasoning for multi-batch emerging
  entities.
\newblock In \emph{Proceedings of the 31st ACM International Conference on
  Information and Knowledge Management}, pp.\  335--344, 2022.

\bibitem[Dai et~al.(2021)Dai, Zheng, Luo, Yang, Liu, Sui, and Chang]{invt}
Dai, D., Zheng, H., Luo, F., Yang, P., Liu, T., Sui, Z., and Chang, B.
\newblock Inductively representing out-of-knowledge-graph entities by optimal
  estimation under translational assumptions.
\newblock In \emph{Proceedings of the 6th Workshop on Representation Learning
  for NLP}, pp.\  83--89, 2021.

\bibitem[Daza et~al.(2021)Daza, Cochez, and Groth]{blp}
Daza, D., Cochez, M., and Groth, P.
\newblock Inductive entity representations from text via link prediction.
\newblock In \emph{Proceedings of the Web Conference 2021}, pp.\  798--808,
  2021.

\bibitem[Devlin et~al.(2019)Devlin, Chang, Lee, and Toutanova]{bert}
Devlin, J., Chang, M.-W., Lee, K., and Toutanova, K.
\newblock {BERT}: Pre-training of deep bidirectional transformers for language
  understanding.
\newblock In \emph{Proceedings of the 2019 Conference of the North American
  Chapter of the Association for Computational Linguistics: Human Language
  Technologies}, pp.\  4171--4186, 2019.

\bibitem[Galkin et~al.(2022)Galkin, Denis, Wu, and Hamilton]{npiece}
Galkin, M., Denis, E., Wu, J., and Hamilton, W.~L.
\newblock Node{P}iece: Compositional and parameter-efficient representations of
  large knowledge graphs.
\newblock In \emph{Proceedings of the 10th International Conference on Learning
  Representations}, 2022.

\bibitem[Geng et~al.(2023)Geng, Chen, Pan, Chen, Jiang, Zhang, and Chen]{rmpi}
Geng, Y., Chen, J., Pan, J.~Z., Chen, M., Jiang, S., Zhang, W., and Chen, H.
\newblock Relational message passing for fully inductive knowledge graph
  completion.
\newblock In \emph{Proceedings of the 39th IEEE International Conference on
  Data Engineering}, 2023.

\bibitem[Gesese et~al.(2022)Gesese, Sack, and Alam]{raild}
Gesese, G.~A., Sack, H., and Alam, M.
\newblock {RAILD}: Towards leveraging relation features for inductive link
  prediction in knowledge graphs.
\newblock In \emph{Proceedings of the 11th International Joint Conference on
  Knowledge Graphs}, 2022.

\bibitem[Glorot \& Bengio(2010)Glorot and Bengio]{glor}
Glorot, X. and Bengio, Y.
\newblock Understanding the difficulty of training deep feedforward neural
  networks.
\newblock In \emph{Proceedings of the 13th International Conference on
  Artificial Intelligence and Statistics}, pp.\  249--256, 2010.

\bibitem[Hamaguchi et~al.(2017)Hamaguchi, Oiwa, Shimbo, and Matsumoto]{mean}
Hamaguchi, T., Oiwa, H., Shimbo, M., and Matsumoto, Y.
\newblock Knowledge transfer for out-of-knowledge-base entities: A graph neural
  network approach.
\newblock In \emph{Proceedings of the 26th International Joint Conference on
  Artificial Intelligence}, pp.\  1802--1808, 2017.

\bibitem[Hamilton et~al.(2017)Hamilton, Ying, and Leskovec]{gsage}
Hamilton, W.~L., Ying, R., and Leskovec, J.
\newblock Inductive representation learning on large graphs.
\newblock In \emph{Proceedings of the 31st Conference on Neural Information
  Processing System}, pp.\  1025–1035, 2017.

\bibitem[He et~al.(2016)He, Zhang, Ren, and Sun]{resi}
He, K., Zhang, X., Ren, S., and Sun, J.
\newblock Deep residual learning for image recognition.
\newblock In \emph{2016 IEEE Conference on Computer Vision and Pattern
  Recognition}, pp.\  770--778, 2016.

\bibitem[Hong et~al.(2023)Hong, Kim, Kang, Myaeng, and Whang]{acl}
Hong, G., Kim, J., Kang, J., Myaeng, S.-H., and Whang, J.~J.
\newblock Discern and answer: Mitigating the impact of misinformation in
  retrieval-augmented models with discriminators.
\newblock \emph{arXiv preprint arXiv:2305.01579}, 2023.
\newblock \doi{10.48550/arXiv.2305.01579}.

\bibitem[Ji et~al.(2022)Ji, Pan, Cambria, Marttinen, and Yu]{survey}
Ji, S., Pan, S., Cambria, E., Marttinen, P., and Yu, P.~S.
\newblock A survey on knowledge graphs: Representation, acquisition, and
  applications.
\newblock \emph{IEEE Transactions on Neural Networks and Learning Systems},
  33\penalty0 (2):\penalty0 494--514, 2022.

\bibitem[Jin et~al.(2022)Jin, Wang, Du, Zhang, Zhang, Wipf, Yu, and Gan]{angel}
Jin, J., Wang, Y., Du, K., Zhang, W., Zhang, Z., Wipf, D., Yu, Y., and Gan, Q.
\newblock Inductive relation prediction using analogy subgraph embeddings.
\newblock In \emph{Proceedings of the 10th International Conference on Learning
  Representations}, 2022.

\bibitem[Kipf \& Welling(2017)Kipf and Welling]{gcn}
Kipf, T.~N. and Welling, M.
\newblock Semi-supervised classification with graph convolutional networks.
\newblock In \emph{Proceedings of the 5th International Conference on Learning
  Representations}, 2017.

\bibitem[Kwak et~al.(2022)Kwak, Lee, Whang, and Jo]{robokg}
Kwak, J.~H., Lee, J., Whang, J.~J., and Jo, S.
\newblock Semantic grasping via a knowledge graph of robotic manipulation: A
  graph representation learning approach.
\newblock \emph{IEEE Robotics and Automation Letters}, 7\penalty0 (4):\penalty0
  9397--9404, 2022.

\bibitem[Lin et~al.(2022)Lin, Liu, Xu, Pan, Zhu, Zhang, and Zhao]{conglr}
Lin, Q., Liu, J., Xu, F., Pan, Y., Zhu, Y., Zhang, L., and Zhao, T.
\newblock Incorporating context graph with logical reasoning for inductive
  relation prediction.
\newblock In \emph{Proceedings of the 45th International ACM SIGIR Conference
  on Research and Development in Information Retrieval}, pp.\  893--903, 2022.

\bibitem[Liu et~al.(2017)Liu, Wu, and Yang]{anal}
Liu, H., Wu, Y., and Yang, Y.
\newblock Analogical inference for multi-relational embeddings.
\newblock In \emph{Proceedings of the 37th International Conference on Machine
  Learning}, pp.\  2168--2178, 2017.

\bibitem[Liu et~al.(2021)Liu, Grau, Horrocks, and Kostylev]{indigo}
Liu, S., Grau, B., Horrocks, I., and Kostylev, E.
\newblock {INDIGO}: {GNN}-based inductive knowledge graph completion using
  pair-wise encoding.
\newblock In \emph{Proceedings of the 35th Conference on Neural Information
  Processing Systems}, pp.\  2034--2045, 2021.

\bibitem[Maas et~al.(2013)Maas, Hannun, and Ng]{leaky}
Maas, A.~L., Hannun, A.~Y., and Ng, A.~Y.
\newblock Rectifier nonlinearities improve neural network acoustic models.
\newblock In \emph{ICML 2013 Workshop on Deep Learning for Audio, Speech and
  Language Processing}, 2013.

\bibitem[Mai et~al.(2021)Mai, Zheng, Yang, and Hu]{compile}
Mai, S., Zheng, S., Yang, Y., and Hu, H.
\newblock Communicative message passing for inductive relation reasoning.
\newblock In \emph{Proceedings of the 35th AAAI Conference on Artificial
  Intelligence}, pp.\  4294--4302, 2021.

\bibitem[Markowitz et~al.(2022)Markowitz, Balasubramanian, Mirtaheri,
  Annavaram, Galstyan, and Steeg]{statik}
Markowitz, E., Balasubramanian, K., Mirtaheri, M., Annavaram, M., Galstyan, A.,
  and Steeg, G.~V.
\newblock {StATIK}: Structure and text for inductive knowledge graph
  completion.
\newblock In \emph{Findings of the Association for Computational Linguistics:
  NAACL 2022}, pp.\  604--615, 2022.

\bibitem[Nathani et~al.(2019)Nathani, Chauhan, Sharma, and Kaul]{kbat}
Nathani, D., Chauhan, J., Sharma, C., and Kaul, M.
\newblock Learning attention-based embeddings for relation prediction in
  knowledge graphs.
\newblock In \emph{Proceedings of the 57th Annual Meeting of the Association
  for Computational Linguistics}, pp.\  4710--4723, 2019.

\bibitem[Sadeghian et~al.(2019)Sadeghian, Armandpour, patrick Ding, and
  Wang]{drum}
Sadeghian, A., Armandpour, M., patrick Ding, and Wang, D.~Z.
\newblock {DRUM}: End-to-end differentiable rule mining on knowledge graphs.
\newblock In \emph{Proceedings of the 33rd Conference on Neural Information
  Processing Systems}, pp.\  15347--15357, 2019.

\bibitem[Sato et~al.(2021)Sato, Yamada, and Kashima]{rgin}
Sato, R., Yamada, M., and Kashima, H.
\newblock Random features strengthen graph neural networks.
\newblock In \emph{Proceedings of the 2021 SIAM International Conference on
  Data Mining}, pp.\  333--341, 2021.

\bibitem[Schlichtkrull et~al.(2018)Schlichtkrull, Kipf, Bloem, van~den Berg,
  Titov, and Welling]{rgcn}
Schlichtkrull, M., Kipf, T.~N., Bloem, P., van~den Berg, R., Titov, I., and
  Welling, M.
\newblock Modeling relational data with graph convolutional networks.
\newblock In \emph{Proceedings of the 15th International Semantic Web
  Conference}, pp.\  593--607, 2018.

\bibitem[Sun et~al.(2019)Sun, Deng, Nie, and Tang]{rotate}
Sun, Z., Deng, Z.-H., Nie, J.-Y., and Tang, J.
\newblock {RotatE}: Knowledge graph embedding by relational rotation in complex
  space.
\newblock In \emph{Proceedings of the 7th International Conference on Learning
  Representations}, 2019.

\bibitem[Teru et~al.(2020)Teru, Denis, and Hamilton]{grail}
Teru, K., Denis, E., and Hamilton, W.
\newblock Inductive relation prediction by subgraph reasoning.
\newblock In \emph{Proceedings of the 37th International Conference on Machine
  Learning}, pp.\  9448--9457, 2020.

\bibitem[Toutanova \& Chen(2015)Toutanova and Chen]{fb}
Toutanova, K. and Chen, D.
\newblock Observed versus latent features for knowledge base and text
  inference.
\newblock In \emph{Proceedings of the 3rd Workshop on Continuous Vector Space
  Models and their Compositionality}, pp.\  57--66, 2015.

\bibitem[Vashishth et~al.(2020)Vashishth, Sanyal, Nitin, and Talukdar]{cgcn}
Vashishth, S., Sanyal, S., Nitin, V., and Talukdar, P.
\newblock Composition-based multi-relational graph convolutional networks.
\newblock In \emph{Proceedings of the 8th International Conference on Learning
  Representations}, 2020.

\bibitem[Vaswani et~al.(2017)Vaswani, Shazeer, Parmar, Uszkoreit, Jones, Gomez,
  Łukasz Kaiser, and Polosukhin]{att}
Vaswani, A., Shazeer, N., Parmar, N., Uszkoreit, J., Jones, L., Gomez, A.~N.,
  Łukasz Kaiser, and Polosukhin, I.
\newblock Attention is all you need.
\newblock In \emph{Proceedings of the 31st Conference on Neural Information
  Processing Systems}, pp.\  5998--6008, 2017.

\bibitem[Veličković et~al.(2018)Veličković, Cucurull, Casanova, Romero,
  Liò, and Bengio]{gat}
Veličković, P., Cucurull, G., Casanova, A., Romero, A., Liò, P., and Bengio,
  Y.
\newblock Graph attention networks.
\newblock In \emph{Proceedings of the 6th International Conference on Learning
  Representations}, 2018.

\bibitem[Wang et~al.(2022)Wang, Zhou, Pan, Dong, Song, and Sha]{cfag}
Wang, C., Zhou, X., Pan, S., Dong, L., Song, Z., and Sha, Y.
\newblock Exploring relational semantics for inductive knowledge graph
  completion.
\newblock In \emph{Proceedings of the 36th AAAI Conference on Artificial
  Intelligence}, pp.\  4184--4192, 2022.

\bibitem[Wang et~al.(2021)Wang, Ren, and Leskovec]{pcon}
Wang, H., Ren, H., and Leskovec, J.
\newblock Relational message passing for knowledge graph completion.
\newblock In \emph{Proceedings of the 27th ACM SIGKDD Conference on Knowledge
  Discovery and Data Mining}, pp.\  1697--1707, 2021.

\bibitem[Wang et~al.(2019)Wang, Han, Li, and Pan]{lan}
Wang, P., Han, J., Li, C., and Pan, R.
\newblock Logic attention based neighborhood aggregation for inductive
  knowledge graph embedding.
\newblock In \emph{Proceedings of the 33rd AAAI Conference on Artificial
  Intelligence}, pp.\  7152--7159, 2019.

\bibitem[Wang et~al.(2017)Wang, Mao, Wang, and Guo]{kgs}
Wang, Q., Mao, Z., Wang, B., and Guo, L.
\newblock Knowledge graph embedding: A survey of approaches and applications.
\newblock \emph{IEEE Transactions on Knowledge and Data Engineering},
  29\penalty0 (12):\penalty0 2724--2743, 2017.

\bibitem[Xiong et~al.(2017)Xiong, Hoang, and Wang]{nell995}
Xiong, W., Hoang, T., and Wang, W.~Y.
\newblock Deep{P}ath: A reinforcement learning method for knowledge graph
  reasoning.
\newblock In \emph{Proceedings of the 2017 Conference on Empirical Methods in
  Natural Language Processing}, pp.\  564--573, 2017.

\bibitem[Xu et~al.(2019)Xu, Hu, Leskovec, and Jegelka]{gin}
Xu, K., Hu, W., Leskovec, J., and Jegelka, S.
\newblock How powerful are graph neural networks?
\newblock In \emph{Proceedings of the 7th International Conference on Learning
  Representations}, 2019.

\bibitem[Xu et~al.(2022)Xu, Zhang, He, Chao, and Yan]{snri}
Xu, X., Zhang, P., He, Y., Chao, C., and Yan, C.
\newblock Subgraph neighboring relations infomax for inductive link prediction
  on knowledge graphs.
\newblock In \emph{Proceedings of the 31st International Joint Conference on
  Artificial Intelligence}, pp.\  2341--2347, 2022.

\bibitem[Yan et~al.(2022)Yan, Ma, Gao, Tang, and Chen]{cbgnn}
Yan, Z., Ma, T., Gao, L., Tang, Z., and Chen, C.
\newblock Cycle representation learning for inductive relation prediction.
\newblock In \emph{Proceedings of the 39th International Conference on Machine
  Learning}, pp.\  24895--24910, 2022.

\bibitem[Yang et~al.(2015)Yang, tau Yih, He, Gao, and Deng]{distm}
Yang, B., tau Yih, W., He, X., Gao, J., and Deng, L.
\newblock Embedding entities and relations for learning and inference in
  knowledge bases.
\newblock In \emph{Proceedings of the 3rd International Conference on Learning
  Representations}, 2015.

\bibitem[Yang et~al.(2017)Yang, Yang, and Cohen]{nelp}
Yang, F., Yang, Z., and Cohen, W.~W.
\newblock Differentiable learning of logical rules for knowledge base
  reasoning.
\newblock In \emph{Proceedings of the 31st Conference on Neural Information
  Processing Systems}, pp.\  2319--2328, 2017.

\bibitem[Ying et~al.(2021)Ying, Cai, Luo, Zheng, Ke, He, Shen, and Liu]{gmer}
Ying, C., Cai, T., Luo, S., Zheng, S., Ke, G., He, D., Shen, Y., and Liu, T.-Y.
\newblock Do transformers really perform badly for graph representation?
\newblock In \emph{Proceedings of the 35th Conference on Neural Information
  Processing Systems}, pp.\  28877--28888, 2021.

\bibitem[Zha et~al.(2022)Zha, Chen, and Yan]{bertrl}
Zha, H., Chen, Z., and Yan, X.
\newblock Inductive relation prediction by {BERT}.
\newblock In \emph{Proceedings of the 36th AAAI Conference on Artificial
  Intelligence}, pp.\  5923--5931, 2022.

\bibitem[Zhang \& Yao(2022)Zhang and Yao]{red}
Zhang, Y. and Yao, Q.
\newblock Knowledge graph reasoning with relational digraph.
\newblock In \emph{Proceedings of the ACM Web Conference 2022}, pp.\  912--924,
  2022.

\bibitem[Zhu et~al.(2021)Zhu, Zhang, Xhonneux, and Tang]{nbf}
Zhu, Z., Zhang, Z., Xhonneux, L.-P., and Tang, J.
\newblock Neural {B}ellman-{F}ord networks: A general graph neural network
  framework for link prediction.
\newblock In \emph{Proceedings of the 35th Conference on Neural Information
  Processing System}, pp.\  29476--29490, 2021.

\end{thebibliography}
\bibliographystyle{icml2023}

\newpage
\appendix
\onecolumn
\section{Generating Datasets for Inductive Knowledge Graph Completion}
\label{app:data}

\begin{algorithm}
\small
\caption{Generating Datasets for Inductive Knowledge Graph Completion}
\label{alg:new}
\begin{flushleft}
\textbf{Input}: $\widetilde{G} = (\sV, \sR, \sE)$, $n_{\text{tr}}$, $n_{\text{inf}}$, $p_{\text{rel}}$, $p_{\text{tri}}$\\
\textbf{Output}: $\widetilde{G_{\text{tr}}}=(\sV_{\text{tr}},\sR_{\text{tr}},\sE_{\text{tr}})$ and $\widetilde{G_{\text{inf}}}=(\sV_{\text{inf}},\sR_{\text{inf}},\sE_{\text{inf}})$
\end{flushleft}
\begin{algorithmic}[1]
\STATE $\widetilde{G}\leftarrow$ Giant connected component of $\widetilde{G}$.
\STATE Randomly split $\sR$ into $\sR_{\text{tr}}$ and $\sR_{\text{inf}}$ such that $|\sR_{\text{tr}}|:|\sR_{\text{inf}}|=(1-p_{\text{rel}}):p_{\text{rel}}$.
\STATE Uniformly sample $n_{\text{tr}}$ entities from $\sV$ and form $\sV_{\text{tr}}$ by taking the sampled entities and their two-hop neighbors. We select at most 50 neighbors per entity for each hop to prevent exponential growth.
\STATE $\sE_{\text{tr}}\coloneqq\{(v_i,r,v_j)|v_i\in\sV_{\text{tr}}, v_j\in\sV_{\text{tr}}, r\in\sR_{\text{tr}},(v_i, r, v_j) \in \sE\}$.
\STATE $\sE_{\text{tr}} \leftarrow$ Triplets in the giant connected component of $\sE_{\text{tr}}$.
\STATE $\sV_{\text{tr}} \leftarrow$ Entities involved in $\sE_{\text{tr}}$.
\STATE $\sR_{\text{tr}} \leftarrow$ Relations involved in $\sE_{\text{tr}}$.
\STATE Let $\widetilde{G}'$ be the subgraph of $\widetilde{G}$ where the entities in $\sV_{\text{tr}}$ are removed.
\STATE In $\widetilde{G}'$, uniformly sample $n_{\text{inf}}$ entities and form $\sV_{\text{inf}}$ by taking the sampled entities and their two-hop neighbors. We select at most 50 neighbors per entity for each hop to prevent exponential growth.
\STATE $\sE_{\text{inf}}\coloneqq\sX\cup\sY$ such that $|\sX|:|\sY|=(1-p_{\text{tri}}):p_{\text{tri}}$ where $\sX\coloneqq\{(v_i,r,v_j)|v_i\in\sV_{\text{inf}}, v_j\in\sV_{\text{inf}}, r\in\sR_{\text{tr}},(v_i, r, v_j) \in \sE\}$ and $\sY\coloneqq\{(v_i,r,v_j)|v_i\in\sV_{\text{inf}}, v_j\in\sV_{\text{inf}}, r\in\sR_{\text{inf}},(v_i, r, v_j) \in \sE\}$.
\STATE $\sE_{\text{inf}} \leftarrow$ Triplets in the giant connected component of $\sE_{\text{inf}}$.
\STATE $\sV_{\text{inf}} \leftarrow$ Entities involved in $\sE_{\text{inf}}$.
\STATE $\sR_{\text{inf}} \leftarrow$ Relations involved in $\sE_{\text{inf}}$.
\end{algorithmic}
\end{algorithm}

Algorithm~\ref{alg:new} shows how we generate the datasets used in Section~\ref{sec:exp}. Also, Table~\ref{tb:data} and Table~\ref{tb:parad} show the statistic of the datasets and the hyperparameters used to create the datasets, respectively. 

As described in Section~\ref{sec:problem}, $\sE_{\text{tr}}$ is divided into $\sF_{\text{tr}}$ and $\sT_{\text{tr}}$. How to split and use $\sE_{\text{tr}}$ is a model-dependent design choice. On the other hand, $\sE_{\text{inf}}$ is divided into three pairwise disjoint sets, $\sF_{\text{inf}}$, $\sT_{\text{val}}$, and $\sT_{\text{test}}$ with a ratio of 3:1:1. For a fair comparison, these three sets are fixed, and the same sets are provided to each model.

In Section~\ref{sec:rgraph}, we mentioned that we add reverse relations and triplets. While this addition is essential for GNN-based methods~\cite{cgcn,red}, we should add the reverse relations and triplets after we split $\sF_{\text{inf}}$, $\sT_{\text{val}}$, and $\sT_{\text{test}}$ to prevent data leakage problems. Similarly, we add the reverse relations and triplets after we split $\sF_{\text{tr}}$ and $\sT_{\text{tr}}$.

\section{Details about the Baseline Methods}
\label{app:base}
All experiments were conducted with GeForce RTX 2080 Ti, GeForce RTX 3090 or RTX A6000, depending on the implementations of each method. We modified all the baseline models except \raild and \rmpi so that the models accept new relations since they do not consider new relations at inference time. We used the original implementations provided by the authors of the models with minimal modification (if needed) and used the default setting except for the things described below. 

Since the original implementations of \nelp and \drum include entities in $\sV_{\text{tr}}$ as candidates for a prediction task at inference time, we excluded them from the candidates for a fair comparison. On the other hand, the implementations of \blp and \raild restrict the candidates to be the entities involved in $\sT_\text{test}$; so we modified this module to consider all entities in $\sV_{\text{inf}}$ to be candidates.

\blp and \qblp require pre-trained vectors for entities and \raild requires pre-trained vectors for both entities and relations, where the pre-trained vectors are produced by feeding text descriptions or names into BERT~\cite{bert}. Among our datasets, NL does not have text descriptions, FB has text descriptions only for entities, and WK has text descriptions for both entities and relations. We provided the pre-trained vectors with \blp, \qblp and \raild using the available information.

We tuned \blp with learning rate $\in \{0.00001, 0.00002, 0.00005\}$ and L2 regularization coefficient $\in \{0, 0.001, 0.01\}$. \qblp was tuned with learning rate $\in \{0.0001, 0.0005\}$, the number of transformer layers $\in \{2,3,4\}$ and the number of GCN layers $\in \{2,3\}$. For \grail, \compile and \snri, we set the early stop patience to be 10 validation trials and the number of total epochs to be 10. Following the original setting of \rmpi, we used the Schema Enhanced \rmpi for \nla, \nlb, \nlc, and \nld, and used the Randomly Initialized \rmpi for \nlz and \nlg. We tuned \red with weight decay $\in \{0.00001, 0.01\}$, dropout rate $\in \{0, 0.3\}$ and the number of layers $\in \{3,4\}$. We set the early stop patience to be 10 epochs.

Since the implementation of \raild does not contain the codes for obtaining node2vec representations of relations, we used the official C++ implementation of node2vec\footnote{\url{https://github.com/snap-stanford/snap/tree/master/examples/node2vec}} to calculate the representations of relations.

The original implementation of \cgcn can only be applied to the transductive setting for both entities and relations. We modified \cgcn so that the model also uses randomly initialized embeddings for new entities appearing at inference time. \cgcn is tuned with the number of GCN layers $\in \{1,2,3\}$, learning rate $\in \{0.0001, 0.001\}$ and the number of bases $\in \{-1,20,40\}$, where -1 denotes the case where each relation has its own embedding. We tuned NodePiece with the size of relational context $\in \{4,12\}$ and the margin $\in \{15, 20, 25\}$.

\paragraph*{Missing Baselines.} We could not include \pcon\cite{pcon} and \tact\cite{tact} in our experiments since their original source codes were written only for relation prediction but not for link prediction. \conglr~\cite{conglr} and \cbgnn~\cite{cbgnn} sample 50 negative candidates for each query, following the experimental setting of \grail. Unlike \grail, \compile and \snri, the original implementations of \conglr and \cbgnn do not provide the code for expanding the candidate set to $\sV_\text{inf}$, so we could not include them as baseline methods. Since the entity sets of training and inference graphs are disjoint in our setting, we could not include baselines assuming new entities should be attached to known entities, such as \mean~\cite{mean} and \lan~\cite{lan}. We could not include \angel~\cite{angel} in our experiments because the results in \cite{angel} are not reproducible.

\section{Hyperparameters of \ours}
\label{app:hyper}
For \ours, we performed validation every 200 epochs for a total of 10,000 epochs. We tuned \ours with 10 negative samples, $d'\in \{32, 64, 128, 256\}$, $\widehat{d}'\in \{128, 256\}$, $L \in \{1,2,3\}$, $\widehat{L} \in \{2,3,4\}$, $K\in \{8, 16\}$, $\widehat{K}\in \{8, 16\}$, $\gamma \in \{1.0, 1.5, 2.0, 2.5\}$, $B\in \{1, 5, 10\}$ and the learning rate $\in \{0.0005, 0.001\}$. We observed that the best performance of \ours is achieved when $B=10$, showing the effectiveness of our binning strategy used in (\ref{eq:indx}) described in Section~\ref{sec:rel}.

\begin{table}[hb]
\small
\centering
\caption{Datasets for Inductive Knowledge Graph Completion.}
\label{tb:data}
\setlength{\tabcolsep}{0.59em}
\begin{tabular}{ccccccccccccc}
\toprule
 & \multicolumn{3}{c}{\nld} & \multicolumn{3}{c}{\nlc} & \multicolumn{3}{c}{\nlb} & \multicolumn{3}{c}{\nla} \\
 & $|\sV|$ & $|\sR|$ & $|\sE|$ & $|\sV|$ & $|\sR|$ & $|\sE|$ & $|\sV|$ & $|\sR|$ & $|\sE|$ & $|\sV|$ & $|\sR|$ & $|\sE|$ \\
\midrule
$\widetilde{G_\text{tr}}$ & 1,258 & 55 & 7,832 & 2,607 & 96 & 11,058 & 4,396 & 106 & 17,578 & 4,396 & 106 & 17,578 \\
$\widetilde{G_\text{inf}}$ & 1,709 & 53 & 3,964 & 1,578 & 116 & 3,031 & 2,335 & 119 & 4,294 & 2,146 & 120 & 3,717 \\
\midrule
 & \multicolumn{3}{c}{\wkd} & \multicolumn{3}{c}{\wkc} & \multicolumn{3}{c}{\wkb} & \multicolumn{3}{c}{\wka} \\
 & $|\sV|$ & $|\sR|$ & $|\sE|$ & $|\sV|$ & $|\sR|$ & $|\sE|$ & $|\sV|$ & $|\sR|$ & $|\sE|$ & $|\sV|$ & $|\sR|$ & $|\sE|$ \\
\midrule
$\widetilde{G_\text{tr}}$ & 9,784 & 67 & 49,875 & 6,853 & 52 & 28,741 & 12,022 & 72 & 82,481 & 12,659 & 47 & 41,873 \\
$\widetilde{G_\text{inf}}$ & 12,136 & 37 & 22,479 & 2,722 & 65 & 5,717 & 9,328 & 93 & 16,121 & 3,228 & 74 & 5,652 \\
\midrule
 & \multicolumn{3}{c}{\fbd} & \multicolumn{3}{c}{\fbc} & \multicolumn{3}{c}{\fbb} & \multicolumn{3}{c}{\fba} \\
 & $|\sV|$ & $|\sR|$ & $|\sE|$ & $|\sV|$ & $|\sR|$ & $|\sE|$ & $|\sV|$ & $|\sR|$ & $|\sE|$ & $|\sV|$ & $|\sR|$ & $|\sE|$ \\
\midrule
$\widetilde{G_\text{tr}}$ & 4,659 & 134 & 62,809 & 4,659 & 134 & 62,809 & 5,190 & 153 & 85,375 & 5,190 & 163 & 91,571 \\
$\widetilde{G_\text{inf}}$ & 2,624 & 77 & 11,645 & 2,792 & 186 & 15,528 & 4,445 & 205 & 19,394 & 4,097 & 216 & 28,579 \\
\bottomrule
\end{tabular}
\end{table}

\begin{table}[hb]
\small
\centering
\caption{Hyperparameters Used to Create the Datasets.}
\label{tb:parad}
\setlength{\tabcolsep}{0.57em}
\begin{tabular}{cccccccccccccc}
\toprule
 & \nld & \nlc & \nlb & \nla & \nlz & \wkd & \wkc & \wkb & \wka & \fbd & \fbc & \fbb & \fba \\
\midrule
$n_\text{tr}$ & 15 & 60 & 50 & 50 & 20 & 20 & 20 & 30 & 30 & 10 & 10 & 10 & 10 \\
$n_\text{inf}$ & 80 & 50 & 80 & 80 & 80 & 250 & 15 & 80 & 50 & 20 & 20 & 50 & 50 \\
$p_\text{rel}$ & 0.40 & 0.40 & 0.40 & 0.40 & 0.00 & 0.30 & 0.40 & 0.30 & 0.50 & 0.40 & 0.40 & 0.30 & 0.25 \\
$p_\text{tri}$ & 1.00 & 0.75 & 0.50 & 0.25 & 0.00 & 1.00 & 0.75 & 0.50 & 0.25 & 1.00 & 0.75 & 0.50 & 0.25 \\
\bottomrule
\end{tabular}
\end{table}


\end{document}